%% file: neurips_2026.tex
\documentclass{article}
\usepackage{tcolorbox}

     \PassOptionsToPackage{numbers, compress}{natbib}
\usepackage{wrapfig} % in preamble
\usepackage{caption}
 \usepackage[main, final]{neurips_2026}
\usepackage[utf8]{inputenc} % allow utf-8 input
\usepackage[T1]{fontenc}    % use 8-bit T1 fonts
\usepackage{hyperref}       % hyperlinks
\usepackage{url}            % simple URL typesetting
\usepackage{booktabs}       % professional-quality tables
\usepackage{amsfonts}       % blackboard math symbols
\usepackage{nicefrac}       % compact symbols for 1/2, etc.
\usepackage{microtype}      % microtypography
\usepackage{xcolor}         % colors
\usepackage{graphicx}
\usepackage{multirow}
\usepackage{amsmath}
\usepackage{algorithm}
\usepackage{algpseudocode}
\usepackage{colortbl}
\definecolor{rowgray}{gray}{0.93}

\title{Not All Layers Are Equal in Image-to-Video Transfer}

\author{%
  Thinesh Thiyakesan Ponbagavathi$^{1,2}$ \quad
  Constantin Seibold$^{3}$ \quad
  Alina Roitberg$^{4}$ \\
  $^{1}$University of Stuttgart\\
 $^{2}$Intelligent Assistive Systems Lab, University of Hildesheim \\
  $^{3}$University Hospital Heidelberg, Diagnostic and Interventional Radiology \\
  $^{4}$ Multimodal Machine Learning and Assistive Computing Lab, University of Münster \\
}
\begin{document}

\maketitle

\begin{abstract}

Parameter-efficient image-to-video transfer adapts frozen image foundation models to video via lightweight trainable adapters. Existing methods insert the same adapter at every transformer layer, implicitly assuming all layers require equal temporal modeling. A layer-wise sensitivity analysis across four video-native model families and two pretraining paradigms reveals that this assumption is wrong: temporal sensitivity increases with depth, and the pretraining paradigm determines its growth profile. Vision-only models exhibit a gradual rise, while vision-language models show early suppression followed by late recovery.

This finding motivates \textsc{Strap}, a training-free adapter placement rule that derives a spatial-to-temporal transition boundary from a paradigm-matched video model. Spatial adapters are assigned before this boundary and temporal adapters after it, without modifying the adapter architecture itself. On four fine-grained action recognition benchmarks, \textsc{Strap} improves accuracy by up to 11.8\% over uniform placement, reduces trainable parameters by 29 to 48\%, and lowers inference time by up to 10\%. On standard benchmarks, it matches accuracy with fewer parameters. Code is available at https://github.com/th-nesh/STRAP.

\end{abstract}

\section{Introduction}

\begin{wrapfigure}{r}{0.5\textwidth} % {r} for right side, {0.5\textwidth} for half-page width
    \centering
\vspace{-0.5cm}
     \includegraphics[width=\linewidth]{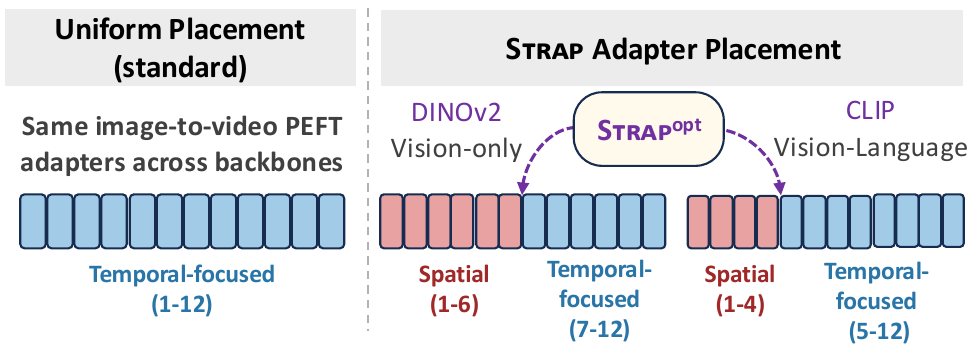}
    \caption{%\textbf{Uniform vs. \textsc{Strap} placement.}
    Existing image-to-video adapters place the same temporal adapter across all layers. \textsc{Strap} instead uses a sensitivity-derived, backbone-specific boundary to place spatial adapters in early layers and temporal adapters in later layers.}
    \label{fig:teaser}
    \vspace{-0.4cm}
\end{wrapfigure}

Adapting frozen image foundation models to video with lightweight temporal modules has become a prominent strategy for video understanding~\cite{St_adaptor,aim,frame2freq,dualpath,m2_clip}, avoiding the substantial data and compute costs of video-native pretraining~\cite{tong2022videomae,wang2024internvideo2scalingfoundationmodels}. Yet while considerable effort has gone into designing these temporal adapters, the question of where to place them has received comparatively little attention: existing methods~\cite{St_adaptor,aim,frame2freq,dualpath,m2_clip} find, through ablations comparing presence versus absence of a single temporal adapter type, that uniform placement performs best within this search space. This leaves open whether placing different adapter types based on layer-wise behavior can outperform uniform placement of temporal adapters.

%Yet while considerable effort has gone into designing these temporal adapters, the question of \emph{where} to place them has received almost no attention: existing methods insert the same adapter architecture uniformly at every layer, implicitly assuming that all layers require the same degree of temporal processing.

% Vision Foundation Models (VFMs) such as CLIP \cite{CLIP} and DINOv2 \cite{dinov2}, learn strong visual representations through large-scale image pretraining.  Building comparable foundation models \cite{tong2022videomae,wang2024internvideo2scalingfoundationmodels} for video requires substantially more data and compute. This has motivated a growing body of work \cite{St_adaptor,aim,frame2freq,dualpath,m2_clip}, which keeps the image-pretrained models frozen and inserts small trainable adapter modules for video understanding. 
% However, these modules are typically placed uniformly across layers, implicitly assuming every layer needs the same temporal adaptation. We challenge this assumption with a simple question: \emph{Do all layers need temporal adaptation?} 
% \red{\emph{Are all layers equal?} }
%\emph{Do all layers need time?} 
%\emph{Do all layers need temporal adaptation?} 

In video-native transformers, there is no reason to expect temporal processing to distribute uniformly across depth. If it does not, then uniform adapter placement \cite{St_adaptor,aim, frame2freq,dualpath,m2_clip} in image-to-video transfer may be misallocating capacity, and might thus be adding temporal modules where they are unnecessary and potentially underserving layers where temporal reasoning concentrates.

We investigate this by analyzing how temporal sensitivity develops across layers in video-native transformers spanning vision-only \cite{tong2022videomae,vjepa2_1} and vision-language \cite{viclip,wang2024internvideo2scalingfoundationmodels} pretraining paradigms. For each model and layer, we apply complementary spatial and temporal perturbations to the input and measure the relative dominance of temporal over spatial sensitivity in the resulting representations.  This yields a depth-wise temporal sensitivity ratio that reveals two findings. First, temporal sensitivity is not uniform: it increases with depth across all model families. Second, the shape of this increase depends on the pretraining paradigm. Vision-only models exhibit a gradual rise across depth, whereas vision-language models show an early suppression of temporal sensitivity in the first layers, followed by a recovery in deeper layers. 
% Rather than transferring video-model features into image models, we use these structural insights to ask a different design question: \emph{where is temporal adaptation actually needed?}

% To answer this, we use video-native transformers as reference models: unlike image-pretrained models, they already model temporal structure, allowing us to study where temporal sensitivity naturally emerges. Our analysis shows that temporal sensitivity is not uniform across layers: it grows with depth, and its transition pattern depends on the pretraining paradigm. Rather than transferring video-model features into image models, we use this structural insight to guide adapter placement. This shifts the design problem from choosing \emph{what} temporal adapter to use to deciding \emph{where} temporal adaptation is actually needed.

% temporal sensitivity emerges gradually rather than uniformly: early layers mostly capture appearance, while later layers become more sensitive to motion and temporal order. Second, these transition patterns differ across pretraining paradigms. 

%\begin{figure}
 %   \centering
  %  \includegraphics[width=0.9\linewidth]{Teaser_simple_version.pdf}
   % \caption{\textbf{Uniform vs. \textsc{Strap} placement.}
    %Existing image-to-video adapters place the same temporal adapter across all layers. \textsc{Strap} instead uses a sensitivity-derived, backbone-specific boundary to place spatial adapters in early layers and temporal adapters in later layers.}
    %\label{fig:teaser}
%\end{figure}

Building on this analysis, we propose \textsc{Strap} (\textbf{S}ensi\textbf{T}ivity-guided Laye\textbf{R}-wise \textbf{A}dapter \textbf{P}lacement), a temporal sensitivity-guided method for placing adapters in image-to-video transfer. \textsc{Strap} uses a frozen video-native model of the same pretraining paradigm as the image backbone to estimate where temporal sensitivity changes most rapidly. This point defines a backbone-specific transition boundary. In the image model, spatial adapters are placed before the boundary, and temporal adapters after it. Importantly, \textsc{Strap} does not introduce a new adapter architecture: it keeps existing adapters unchanged and replaces only their uniform layer assignment with a training-free, sensitivity-guided placement rule.

Our contributions can be summarized as follows:

    ~~(1) A layer-wise temporal sensitivity analysis across four video-native model families and two pretraining paradigms, revealing that temporal sensitivity increases with depth and follows paradigm-specific transition profiles.
    
    ~~(2)  \textsc{Strap}, a training-free placement rule that uses a paradigm-matched video model to determine where spatial and temporal adapters should be assigned in an image backbone.
    
    ~~(3) Evaluation on four fine-grained and two standard action recognition benchmarks. \textsc{Strap} improves fine-grained recognition, where temporal order is critical, while matching standard-benchmark performance, in both cases with fewer trainable parameters than uniform placement. 

% (1) We introduce a layer-wise temporal sensitivity analysis of video foundation models as a basis for adapter placement in image-to-video transfer. Across four model families spanning two pretraining paradigms, this analysis reveals that temporal sensitivity grows with depth and follows paradigm-specific emergence profiles.
% (2) Building on this insight, we propose \textsc{Strap}, a training-free method that uses a frozen paradigm-matched video model to decide where to place spatial adapters and temporal adapters.
% (3) Across four fine-grained and two standard action-recognition benchmarks, \textsc{Strap} achieves strong gains in fine-grained recognition with fewer trainable parameters, while matching or slightly improving standard benchmark performance with reduced parameters. Together, these results show that \emph{where} adapters are placed matters as much as \emph{what} adapters are used.

% (3) Through extensive experiments across six benchmarks: four fine-grained and two standard action-recognition datasets. On fine-grained tasks, \textsc{Strap} improves over uniform placement while using fewer trainable parameters and is competitive with or better than PEFT and fully fine-tuned video baselines. On standard benchmarks, it matches or slightly improves accuracy while reducing parameters. Together, these results show that \emph{where} adapters are placed matters as much as \emph{what} adapters are used. 

\vspace{-0.5em}
\section{Related Work}
\vspace{-0.5em}
\noindent\textbf{Parameter-efficient image-to-video adaptation.}
Parameter-efficient fine-tuning (PEFT) adapts frozen image foundation models to video via lightweight trainable modules while preserving the pretrained backbone. Existing approaches differ in how temporal information is injected. Adapter-based methods, including ST-Adapter~\cite{St_adaptor}, AIM~\cite{aim}, M2-CLIP~\cite{m2_clip}, and Omni-CLIP~\cite{omniclip}, model temporal dependencies via convolutional, temporal difference, or attention-based modules. Extensions such as Frame2Freq~\cite{frame2freq} capture motion in the frequency domain, while DualPath~\cite{dualpath} separates spatial and temporal processing into parallel branches with late fusion. Prompt-based methods, such as VitaCLIP~\cite{VitaCLIP}, adapt CLIP via multimodal prompt learning using learnable context tokens. Fully fine-tuned approaches, including STAN~\cite{STAN}, BIKE~\cite{BIKE}, and ILA~\cite{ILA}, improve temporal reasoning via auxiliary networks and cross-modal alignment, but operate outside the parameter-efficient regime that \textsc{Strap} targets. Despite their architectural diversity, these methods share a common limitation: temporal adaptation is applied uniformly across layers, implicitly assuming all layers require equal temporal modelling. \textsc{Strap} challenges this assumption by deriving backbone-specific adapter placement from layer-wise temporal sensitivity profiles of video-native models.

\noindent\textbf{Layer-wise representation analysis.}
Prior work has performed representation analysis across depth within specific video architectures: VTCD~\cite{VTCD} discovers object-tracking concepts in later layers of video models, while \cite{causality_matters} traces temporal understanding in video-language models to order-sensitive attention pathways.
DiffTrack~\cite{DiffTrack} analyzes temporal correspondence in video diffusion transformers, while related work uses internal representations for controlled generation and motion consistency~\cite{moft,track4gen}.
 These works do not quantify how temporal sensitivity evolves relative to spatial sensitivity across depth, or how this profile varies across pretraining paradigms. \textsc{Strap} uses this depth-wise sensitivity profile as a training-free signal for adapter placement in image-to-video PEFT.

\noindent\textbf{Sensitivity-guided adaptation.}
AdaLoRA~\cite{adalora} adaptively allocates LoRA rank budgets across weight matrices using importance scores estimated during fine-tuning.
Related work applies layer-wise sensitivity signals to quantization~\cite{lsgquant} and layer-adaptive temporal correction in long video generation~\cite{freelunchlongvideogeneration}. These methods target specific downstream tasks and rely on task-specific signals. In contrast, \textsc{Strap} derives a layer-wise sensitivity profile from a frozen, paradigm-matched video model and uses it as a training-free signal for adapter placement across designs, backbones, and datasets — without labels, gradients, or architectural modification.

% On the adaptation side, AdaLoRA~\cite{adalora} demonstrates that allocating parameters based on layer importance outperforms uniform distribution. Related work applies sensitivity analysis to parameter allocation in quantization~\cite{lsgquant}, temporal correction in long video generation~\cite{freelunchlongvideogeneration}, and layer-wise intervention in video-language models~\cite{circuitprobe}. However, these methods either rely on training signals or focus on downstream tasks and do not translate the representation structure into adaptation strategies for image-to-video transfer. \textsc{Strap} bridges this gap by using the sensitivity profile of a video-native model as a blueprint for adapter placement in an image backbone - the first method to derive PEFT placement directly from video-native model behaviour, requiring no labeled data, gradient computation, or architectural modification.

% \section{Preliminaries}

\input{sections/03_temporal_sensitivity_analysis}

\input{sections/04_methods}

\input{experiments}
\vspace{-0.5em}
\section{Limitations}
\label{sec:Limitations}
\textsc{Strap} rests on the assumption that the depth-wise temporal sensitivity profile of a video-native model is a meaningful prior for adapter placement in an image-pretrained backbone of the same pretraining paradigm. While the empirical results support this, a theoretical basis for why paradigm matching is sufficient or when it might fail is missing. The perturbation-based sensitivity metric mixes spatial and temporal perturbations, which differ in semantic severity, and although min-max normalization reduces scale differences, the absolute values of $R(l)$ should be interpreted as relative depth-wise trends rather than calibrated measurements. Finally, \textsc{Strap} assumes a single hard boundary between spatial and temporal adapters. A soft or graduated transition could capture finer structure in the sensitivity profile, particularly for architectures where the transition may not be as sharp.
\vspace{-0.5em}
\section{Conclusion}
This work studied an overlooked question in image-to-video transfer: \emph{do all layers in an image-to-video model need the same temporal adaptation?} The answer is \emph{no}. Temporal sensitivity increases with depth, and the profile of that increase is shaped by how the model was pretrained. \textsc{Strap} uses this observation as a placement rule: spatial adapters go in early layers, temporal adapters in later ones, with the boundary read off a frozen video-native proxy. No new adapter architecture is needed, and no labels or gradients are involved in determining the placement.
The practical payoff is clearest on fine-grained benchmarks, where temporal order separates otherwise similar classes. On standard benchmarks, the accuracy gains are smaller, but the parameter and compute savings are consistent. Whether similar transitions appear in video generation or cross-modal settings is a natural next question. More generally, existing work on image-to-video PEFT has focused almost entirely on designing better adapters. The results suggest that where those adapters are placed matters too, and that this direction has been largely unexplored.

% \textsc{Strap} turns this into a training-free placement rule: spatial adapters early, temporal adapters late, with the boundary derived from a paradigm-matched video proxy- requiring no new architecture, no labels, and no gradients.
% Across backbones, adapter designs, and recognition settings, this structured placement improves the accuracy--efficiency trade-off, with the largest gains when temporal order is most discriminative.
% Our results suggest that in image-to-video PEFT, \textit{where} adapters are placed can be as important as \textit{what} adapters are used.

\begin{ack}
This research was partially  supported by the Deutsche
Forschungsgemeinschaft (DFG) under Germany’s Excellence Strategy – EXC 2120/1 –
390831618 and the Baden-Württemberg Stiftung (Elite Postdoc Program, PRIDRIVE project). This research was also supported by institutional funding of the University of Stuttgart, the University of Hildesheim and the University of Münster.
The authors also thank the International Max Planck Research School 
for Intelligent Systems (IMPRS-IS) for supporting Thinesh Thiyakesan 
Ponbagavathi. 
The authors also gratefully acknowledge the computing time provided on the high-
performance computer HoreKa by the National High-Performance Computing Center at
KIT. This center is jointly supported by the Federal Ministry of Education and 
Research and the Ministry of Science, Research and the Arts of Baden-
Württemberg, as part of the National High-Performance Computing (NHR) joint 
funding program (https://www.nhr-verein.de/en/our-partners). 
HoreKa is partly funded by the German Research Foundation (DFG).
\end{ack}

{
\small

\bibliographystyle{unsrt}
%\balance
\bibliography{neurips}
}

%%%%%%%%%%%%%%%%%%%%%%%%%%%%%%%%%%%%%%%%%%%%%%%%%%%%%%%%%%%%
\input{supplemental}

\newpage

\end{document}

%% file: sections/03_temporal_sensitivity_analysis.tex
\section{Where does Temporal Sensitivity Reside?}\label{sec:analysis}
\input{sections/figures/sentivity_analysis_design}

To gain an intuition about optimal adapter placements in image-to-video models, we analyze how temporal sensitivity is distributed across depth in video-native transformers. The resulting profile can then guide adapter placement in image-to-video transfer. The analysis pipeline is summarized in Fig.~\ref{fig:metric}.

\subsection{Temporal Sensitivity Analysis}
\noindent\textbf{Perturbation Design:}
To probe how the layer-wise representation responds to spatial versus temporal structure, we apply two complementary perturbations to a clip $x$ and compare the resulting embeddings at every transformer block: 
 A spatial perturbation $x_s$ (per-frame Gaussian blur, crop-resize, color jitter) disrupts appearance while preserving frame order and a temporal perturbation $x_t$
(frame reversal, block shuffle, frame drop) disrupts ordering and continuity while preserving the appearance of the retained frames. 

While these two perturbations cannot be assumed to be equivalent in terms of semantic severity, we use them as \emph{probes of relative depth-wise responsiveness}, not as absolute measurements of spatial or temporal information. 

\noindent\textbf{Sensitivity Computation:} A frozen video-native foundation model then processes $x$, $x_s$, and $x_t$. At each block $l \in \{1,\ldots,L\}$, we extract mean-pooled patch-token embeddings $f_l(\cdot)$ and define
\begin{equation}
    S_l = 1 - Sim_{cos}\!\bigl(f_l(x_s), f_l(x)\bigr)
    \quad \text{and} \quad
    T_l = 1 - Sim_{cos}\!\bigl(f_l(x_t), f_l(x)\bigr)
\end{equation}
as its spatial and temporal sensitivity, respectively. We average $S_l$ and $T_l$ over clips, then independently min-max normalize (indicated by $*^{norm}$) each response sequence across layers for every model-dataset pair to prevent the raw perturbation magnitude from dominating the ratio while preserving the depth-wise profile induced by each perturbation type. Inspired by relative sensitivity indices in variance-based analysis~\cite{vsensitivity_score}, we then define \emph{Temporal Sensitivity Ratio} of a block as  $R(l) = {T_l^{\text{norm}}}/{(T_l^{\text{norm}} + S_l^{\text{norm}})}$,
% \begin{equation}
%     R(l) = \frac{T_l^{\text{norm}}}{T_l^{\text{norm}} + S_l^{\text{norm}}}.
%     \label{eq:rl}
% \end{equation}
$R(l)\in[0,1]$ measures relative temporal dominance: values near 0 indicate stronger response to spatial perturbations, while values near 1 indicate stronger response to temporal perturbations. 
The trajectory of $R(l)$ across depth thus indicates where representations transition from appearance-sensitive to temporally-sensitive.

\noindent\textbf{Analysis Setup:}
We compute $R(l)$ for four video-native model families spanning vision-only and vision-language pretraining: VideoMAE \cite{tong2022videomae} (masked reconstruction), V-JEPA 2.1 \cite{vjepa2_1} (predictive latent), ViCLIP \cite{viclip} (contrastive vision–language), and InternVideo2 \cite{wang2024internvideo2scalingfoundationmodels} (a hybrid). For each, we evaluate all available Small/Base/Large variants on $500$ clips per model–dataset pair. We average the normalized $R(l)$ profiles across six benchmarks to capture a global depth-wise sensitivity pattern for each proxy model, covering fine-grained, standard, and human-robot interaction recognition.

\input{sections/figures/sensitivity_analysis_results}

\subsection{\emph{Not All Layers Are Equal} in processing temporal information}

The top row of Fig.~\ref{fig:sensitivity_analysis} shows how $R(l)$ behaves with increasing normalized network depth for the four different models.  $R(l)$ displays a clear trend across all settings as early layers respond more strongly to appearance perturbations, whereas deeper layers become increasingly responsive to temporal-order perturbations,  indicating that temporal sensitivity is not uniformly distributed across depth.

\begin{tcolorbox}[
    colback=gray!5,
    colframe=gray!50,
    boxrule=0.5pt,
    arc=3pt,
    left=6pt, right=6pt, top=4pt, bottom=4pt
]
\textbf{Finding 1:} \textit{The temporal sensitivity of a video network increases with depth.
% : early layers are more appearance-responsive, while deeper layers become increasingly sensitive to temporal-order perturbations.
}
\end{tcolorbox}

While $R(l)$ displays the highest values towards the final layers of a network, its growth profile differs across model families. Vision-only models (VideoMAE and V-JEPA) exhibit a gradual increase of $R(l)$ across depth. In contrast, the vision-language models such as ViCLIP show early suppression, where $R(l)$ decreases in the first few layers before recovering later. This trend can also be observed partially in the hybrid case of InternVideo2: despite its masked-reconstruction component, it retains the early suppression characteristic of vision-language pretraining, but recovers with a smoother, more gradual rise than ViCLIP.
Within each paradigm, the profile shapes of $R(l)$ behave similarly across available size variants albeit the absolute sensitivity may vary. 
The growth profile of $R(l)$ is therefore a signature of the pretraining paradigm and is preserved across model scales.

\begin{tcolorbox}[
   colback=gray!5,
    colframe=gray!50,
    boxrule=0.5pt,
    arc=3pt,
    left=6pt, right=6pt, top=4pt, bottom=4pt
]
\textbf{Finding 2:} \textit{The depth-wise growth of $R(l)$ depends on the pretraining paradigm.
% : early layers are more appearance-responsive, while deeper layers become increasingly sensitive to temporal-order perturbations.
}
\end{tcolorbox}

To verify that our findings generalize, we extend the sensitivity analysis to two
additional video-only models (VideoMAEv2~\cite{videomaev2} and V-JEPA1~\cite{v-jepa})
and two video-language models (UMT~\cite{UMT} and InternVideo~\cite{internvideo}).
Both findings hold across these models; see Appendix~\ref{app:add_backbones} for details.
\subsection{Design Implications}
\label{sec:implications}

From these two findings, we can derive design constraints for image-to-video adapters. Finding~1 implies that uniform placement is suboptimal: distributing temporal modules evenly across depth allocates capacity to layers whose representations are predominantly appearance-driven. Finding~2 further implies that no fixed depth-wise rule can transfer across backbones; instead, the proxy used to determine placement must match the pretraining paradigm of the target image model.

%% file: sections/figures/sentivity_analysis_design.tex
\begin{wrapfigure}{r}{0.51\textwidth}
    \centering
    \vspace{-10pt}
    \includegraphics[width=\linewidth]{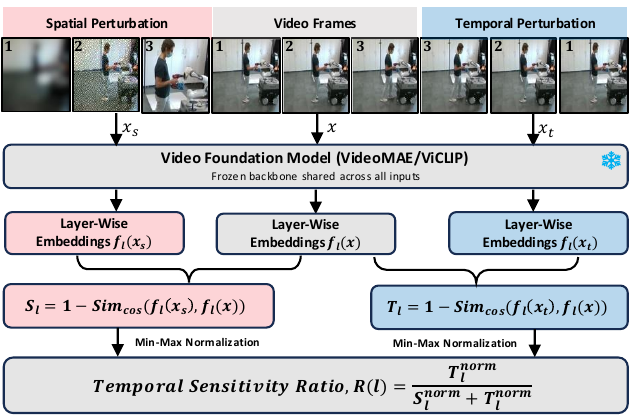}
    \caption{\textbf{Sensitivity Metric Pipeline.} Spatially and temporally perturbed versions of a video clip are passed through a frozen video-native model together with the original clip. Layer-wise cosine distances between mean-pooled patch-token embeddings measure each layer's response to spatial and temporal perturbations, yielding the Temporal Sensitivity Ratio $R(l)$.}
    \label{fig:metric}
    \vspace{-36pt}
\end{wrapfigure}

%% file: sections/figures/sensitivity_analysis_results.tex
\begin{figure}[t]
	\centering
\includegraphics[width=1\textwidth]{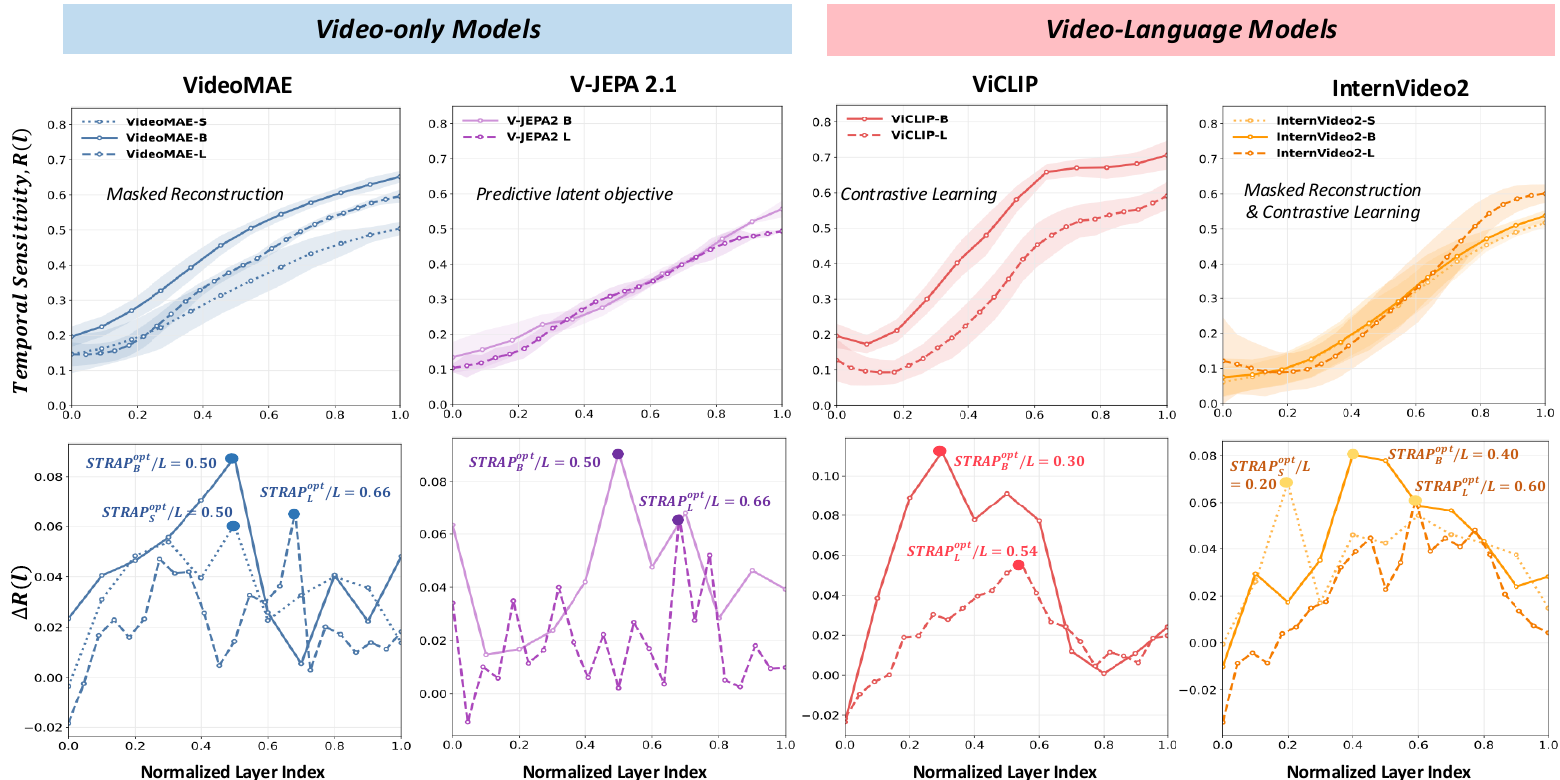}
\vspace{-1em}
    \caption{
\textbf{Temporal sensitivity profiles are depth-dependent and paradigm-determined.} $R(l)$ increases with depth across all models (Finding 1). Vision-only models show gradual emergence; vision-language models exhibit early suppression followed by a late-layer rise (Finding 2), consistent with the influence of cross-modal alignment. Profiles are stable across scales and datasets.}
\vspace{-1em}
	\label{fig:sensitivity_analysis}

\end{figure}

%% file: sections/04_methods.tex
\vspace{-1em}
\section{\textsc{Strap}: Sensitivity-guided Layer-wise Adapter Placement}
\label{sec:strap}
 
Sec.~\ref{sec:analysis} shows that temporal sensitivity emerges non-uniformly across depth and follows a paradigm-dependent profile. \textsc{Strap} (\textbf{S}ensi\textbf{T}ivity-guided Laye\textbf{R}-wise \textbf{A}dapter \textbf{P}lacement) converts this profile into a layer-wise adapter placement rule.

\subsection{Preliminaries}
 
\noindent\textbf{Backbone and adapters.}
We consider a transformer backbone with $L$ layers processing a video $\mathbf{x}=\{x_1,\ldots,x_T\}$. Let $h_l \in \mathbb{R}^{B \times T \times N \times D}$ denote the token representation at layer $l$, where $B$ is the batch size, $T$ the number of frames, $N$ the number of spatial tokens per frame, and $D$ the feature dimension. Adapters are inserted into transformer blocks through residual updates,
\begin{equation}
    h_l' = h_l + A_l(h_l),
\end{equation}
where $A_l(\cdot)$ is a lightweight trainable module and the image-pretrained backbone remains frozen.
 
\noindent\textbf{Spatial adapters.}
Spatial adapters operate independently on each frame and do not introduce cross-frame interactions. Following image-adaptation designs such as AdaptFormer~\cite{adaptformer}, we reshape $h_l$ from $(B,T,N,D)$ to $(B\!\cdot\!T,N,D)$ and apply a bottleneck transformation frame-wise:
\begin{equation}
A_l^{\text{spatial}}(h_l)[t] = f(h_l[t] W_{\text{down}}) W_{\text{up}}, \quad \forall t \in \{1,\dots,T\}
\end{equation}
where $W_{\mathrm{down}}\in\mathbb{R}^{D\times r}$, $W_{\mathrm{up}}\in\mathbb{R}^{r\times D}$, $r\ll D$, and $f$ is a non-linearity. These adapters refine per-frame spatial representations while preserving temporal independence.
 
\noindent\textbf{Temporal adapters.}
Temporal adapters introduce cross-frame interactions into an image-pretrained backbone. We write them generically as
\begin{equation}
A_l^{\text{temporal}}(h_l) = \mathcal{T}_l\!\left(h_l W_{\text{down}}\right) W_{\text{up}}
\end{equation}
where $\mathcal{T}_l$ is a lightweight temporal operator. Different image-to-video PEFT methods instantiate $\mathcal{T}_l$ differently: ST-Adapter~\cite{St_adaptor} uses temporal depth-wise convolutions, Frame2Freq~\cite{frame2freq} applies multi-scale Fourier-domain and temporal convolutions, and M2-CLIP~\cite{m2_clip} combines temporal convolution, spatial convolution, and temporal difference modeling. \textsc{Strap} keeps these temporal adapter designs unchanged and modifies only their layer-wise placement.
 
\subsection{Sensitivity-guided Placement}
\label{sec:placement}

Since the shape of $R(l)$ itself is paradigm-dependent, no fixed layer index applies across backbones and we require a layer-level signal that is comparable across paradigms. We instead consider the discrete derivative
\begin{equation}
\Delta R(l) = R(l+1) - R(l),
\end{equation}
which captures the local rate at which temporal sensitivity emerges across depth. As shown in the bottom row of Fig.~\ref{fig:sensitivity_analysis}, $\Delta R(l)$ admits a clearly identifiable maximum across all model families and scales, despite the differing shapes of $R(l)$ itself. 

We denote the layer at which this peak occurs as \textsc{Strap} optimum
\begin{equation}
\text{\textsc{Strap}}^{\text{opt}} = \arg\max_{l}\, \Delta R(l).
\end{equation}
We interpret $\text{\textsc{Strap}}^{\text{opt}}$ as the depth at which the network most rapidly shifts from appearance-dominated to temporally-dominated representation and use it as the boundary between spatial and temporal adaptation. Layers before this transition are relatively more appearance-responsive, while layers after it become more responsive to temporal-order perturbations. This criterion is analogous to transition or knee-point detection in structured curves~\cite{knedle,raiche}.

\noindent\textbf{Placement rule.}
Given $\text{\textsc{Strap}}^{\text{opt}}$, the adapter types are assigned as:
\begin{equation}
    \text{adapter}(l) = \begin{cases}
        \text{spatial}  & \text{if } l \leq \text{\textsc{Strap}}^{\text{opt}} \\
        \text{temporal} & \text{if } l > \text{\textsc{Strap}}^{\text{opt}}
    \end{cases}
    \label{eq:placement}
\end{equation}
Early layers retain frame-wise spatial adaptation, while later layers receive temporal adapters where the proxy model shows stronger temporal responsiveness.

\noindent\textbf{Backbone matching.}
Each image backbone is paired with a video-native proxy of matching pretraining paradigm and scale. For example, this matches: DINOv2 with VideoMAE, yielding $\text{\textsc{Strap}}^{\text{opt}}=6$ and a 6S/6T split, and CLIP with ViCLIP, yielding $\text{\textsc{Strap}}^{\text{opt}}=4$ and a 4S/8T split. Sweeping the spatial-temporal boundary across configurations confirms that the sensitivity-derived boundary is empirically optimal across datasets (Fig.~\ref{fig:ablation}).

\noindent\textbf{Adapter design.}
\textsc{Strap} is adapter-agnostic: it determines \emph{where} adapters are placed, not \emph{what} adapters are used. We retain the original temporal adapter designs of each base method, ST-Adapter~\cite{St_adaptor}, Frame2Freq~\cite{frame2freq}, and M2-CLIP~\cite{m2_clip}, and replace only their uniform layer assignment with sensitivity-guided placement. Spatial adapters use a lightweight bottleneck adapter following AdaptFormer~\cite{adaptformer}. Prior methods including AIM~\cite{aim} and DualPath~\cite{dualpath} combine spatial and temporal modules but apply the same pattern uniformly across depth.

%% file: experiments.tex
\section{Experiments}
\vspace{-0.5em}
\label{sec:experiments}

\subsection{Experimental Setup}
\label{sec:exp_setup}
\input{table_main_fine_grained}

Unless otherwise, all experiments use frozen DINOv2-B/14~\cite{dinov2} and CLIP-B/16~\cite{CLIP} as the image-pretrained backbones, which we adapt via PEFT. 
We train only the adapter modules and a linear classifier, keeping the image-pretrained backbone fixed. 
Backbone-to-proxy matching and the resulting placement boundaries follow Sec.~\ref{sec:placement}.  We compare \textsc{Strap} against conventional uniform adapter placement under identical training and evaluation settings, using 16-frame inputs at $224{\times}224$ resolution (32 frames for SSv2 and 8 frames for Kinetics) and AdamW for 60 epochs. Additional implementation details are provided in the Appendix \ref{app:implementation}. 

% Motivated by Sec.~\ref{sec:analysis}, where vision-only and vision-language pretrained models show different temporal sensitivity patterns, we match each image backbone to a video reference model with the same pretraining paradigm.
% Thus, DINOv2-B/14 uses the VideoMAE-derived boundary $\text{\textsc{Strap}}^{\text{opt}}=6$, resulting in a 6S/6T spatial-
% temporal split, while CLIP-B/16 uses the ViCLIP-derived boundary $\text{\textsc{Strap}}^{\text{opt}}=4$, resulting in a 4S/8T split.

% We compare \textsc{Strap} against uniform adapter placement under identical training and evaluation settings, using 16-frame inputs at  resolution (32 frames for SSv2) and AdamW for 60 epochs. Additional implementation details are provided in the supplementary material.

% We compare \textsc{Strap} against conventional uniform adapter placement under identical training and evaluation settings.
% All models use 16-frame inputs at $224{\times}224$ resolution, except the ones used for the  SSv2 in standard action recognition, where we use 32 frames. Training is performed with AdamW for 60 epochs. We report Top-1 accuracy, trainable parameter count, and inference time where applicable. 
% Additional implementation details are provided in the supplementary material.

% \subsection{Experimental Setup}
% \label{sec:exp_setup}

%\noindent\textbf{Backbones and adapter families.}

%\noindent\textbf{STRAP placement.}

%\noindent\textbf{Baselines.}

%\noindent\textbf{Implementation details.}

%\noindent\textbf{Evaluation metrics.}

\vspace{-0.5em}
\subsection{Main Results: Fine-Grained Action Recognition}
\label{sec:finegrained_results}

\noindent\textbf{Datasets.} 
We evaluate the main setting on four fine-grained action recognition benchmarks: Diving48~\cite{diving48} (diving competitions), Drive\&Act~\cite{drive_and_act} (driver behavior analysis), IKEA-ASM~\cite{IKEA_ASM} (furniture assembly), and HRI-30~\cite{hri_30} (fine-grained human-robot interaction), where subtle motion and temporal ordering are class-discriminative alongside appearance.

\input{table_symmetric}

\noindent\textbf{Results.} We first compare \textsc{Strap} to uniform adapter placement, our main baseline, for three image-to-video PEFT adapters: ST-Adapter, Frame2Freq, and M2-CLIP. Results are reported for both DINOv2 and CLIP backbones. We also include published fully fine-tuned video models in gray for context, although these methods are outside the parameter-efficient setting targeted by this work.

Table~\ref{tab:finegrained} shows that \textsc{Strap} achieves state-of-the-art results on all four fine-grained benchmarks among the compared methods. Against uniform placement, \textsc{Strap} improves accuracy in most settings while reducing trainable parameters by 29--48\% and inference time by 5--10\%. The largest gains appear on Drive\&Act, IKEA-ASM, and HRI-30: with DINOv2 and ST-Adapter, accuracy improves by +6.7\%, +7.1\%, and +3.7\%, respectively. Gains on Diving48 are smaller. With CLIP, \textsc{Strap} improves ST-Adapter and M2-CLIP on all datasets, while Frame2Freq shows small drops on Diving48 and IKEA-ASM. These results suggest that layer-wise spatial-temporal placement is useful for fine-grained video understanding with fewer trainable parameters needed.

To investigate where \textsc{Strap} helps most, we also use the split into symmetric and non-symmetric classes~\cite{STEP,frame2freq}. %\red{Todo: cite ICRA and CVPR}. 
Symmetric actions, such as \textit{opening} vs. \textit{closing} or \textit{taking} vs. \textit{putting}, share similar movements but differ in temporal direction. 
They are therefore hard to resolve from appearance alone and require modeling how the action unfolds.
Table~\ref{tab:sym_nonsym_wrap} shows a clear trend: \textsc{Strap} brings the largest gains on these temporally ambiguous classes. 
For example, on IKEA-ASM, symmetric-action accuracy improves by +11.8\%, +5.8\%, and +8.4\% depending on the adapter type. 
These gains are usually larger than those on non-symmetric actions, suggesting that \textsc{Strap} allocates temporal capacity where temporal direction matters most.

\subsection{Does Placement Matter Beyond Parameter Count?}
\label{sec:placement_vs_capacity}

\noindent\textbf{Placement direction.}
We next test whether the order of spatial and temporal adapters matters (Table \ref{tab:component_ablation}, upper section). 
Reversing the layout, i.e., placing temporal adapters in early layers and spatial adapters in later layers, strongly degrades performance. 
Compared to the original order, accuracy drops by $-11.1\%$  on Drive\&Act, $-40.7\%$  on IKEA-ASM, and $-15.7\%$  on HRI-30. This shows that the gain is not due to mixing both adapter types alone, and the spatial-to-temporal direction is critical.

\noindent\textbf{Component contribution.}
In Table \ref{tab:component_ablation} (middle section), we isolate spatial and temporal adapters using Frame2Freq with DINOv2 and the \textsc{Strap} boundary $\text{\textsc{Strap}}^{\text{opt}}=6$. 
Spatial-only adaptation is weak, especially on HRI-30. 
In contrast, using temporal adapters only in the later layers already outperforms uniform temporal placement, despite using fewer parameters. 
Our strategy using both performs best, indicating that early spatial adaptation remains useful, but temporal adaptation is better concentrated in later layers.

\noindent\textbf{Matched-capacity comparison.} 
In theory,  the improved performance could be simply because our method uses fewer trainable parameters, thereby reducing overfitting. 
We test this in Table \ref{tab:component_ablation} (bottom section), by comparing against a width-adjusted uniform baseline with the same parameter count as \textsc{Strap}. 
This matched-capacity baseline performs worse, with drops of $-13.3$ points on Drive\&Act, $-3.7$ points on IKEA-ASM, and $-3.5$ points on HRI-30. 
Thus, the gains are not explained by parameter count alone, but by the layer-wise placement of spatial and temporal adapters.

\subsection{Analyzing the Spatial-Temporal Adapter Boundary}
\label{sec:boundary_ablation}

\begin{figure}[t]
	\centering
\includegraphics[width=\textwidth, trim={0.4cm 0.45cm 0.4cm 1.2cm},clip]{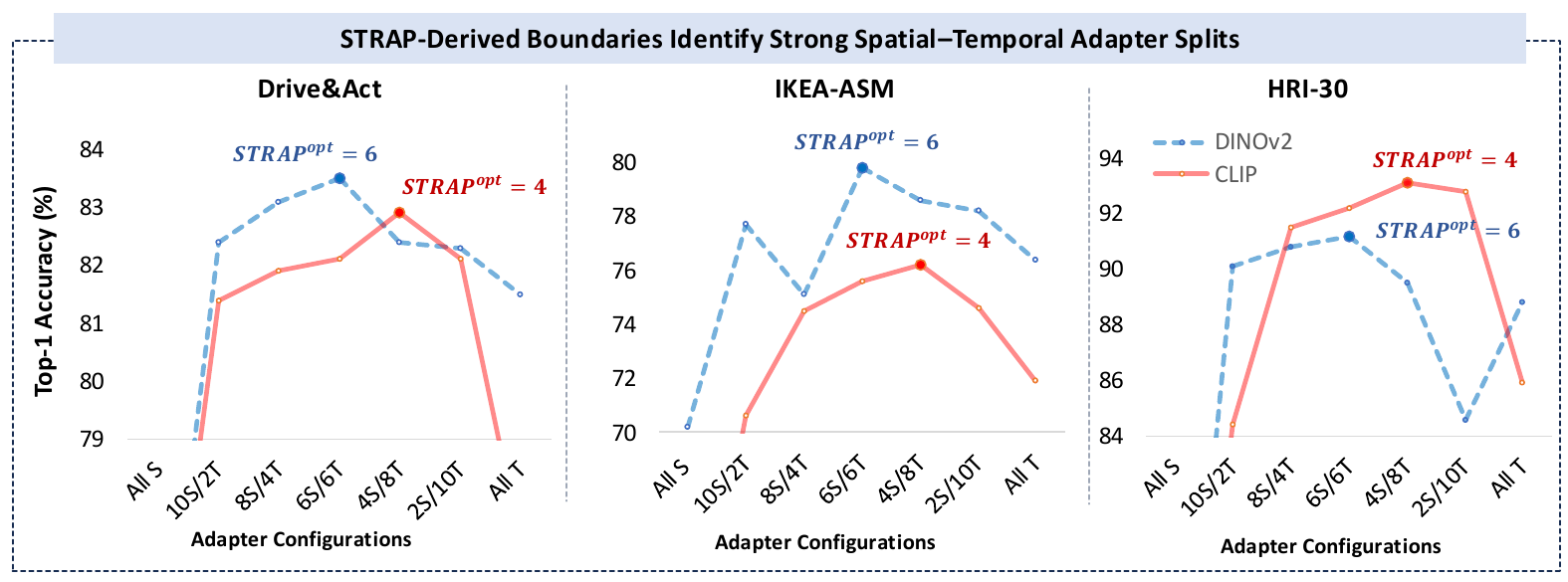}

    \caption{
Ablation of spatial-temporal adapter allocation across layers. We vary the number of spatial~(S) and temporal~(T) adapters under varying depth.% (parameter count varies accordingly).
 %Balanced allocations (shaded) consistently outperform spatial- or temporal-heavy configurations across diverse datasets.
 }
\vspace{-1em}
	\label{fig:ablation}

\end{figure}

\input{ablation_tables}

We now investigate how the spatial-temporal adapter boundary affects performance.
In Figure~\ref{fig:ablation}, we vary the spatial/temporal adapter split while keeping the number of adapted layers fixed for DINOv2 and CLIP.
Balanced allocations generally outperform spatial- or temporal-heavy configurations, with the best results lying near the sensitivity-derived boundaries: $\text{\textsc{Strap}}^{\text{opt}}=6$ for DINOv2 and $\text{\textsc{Strap}}^{\text{opt}}=4$ for CLIP.
% These boundaries follow the sensitivity analysis in Section~\ref{sec:analysis}: the vision-only reference suggests a later transition point for DINOv2, while the vision-language reference suggests an earlier transition point for CLIP.
Using the boundary induced by the opposite proxy, i.e., $\text{\textsc{Strap}}^{\text{opt}}=4$ for DINOv2 or $\text{\textsc{Strap}}^{\text{opt}}=6$ for CLIP, moves the allocation away from the empirical peak.
This acts as a proxy-mismatch diagnostic and suggests that the transition point is paradigm-specific rather than directly transferable across backbones.
Thus, \textsc{Strap} provides a training-free boundary selection rule without labeled validation or split-specific tuning, while remaining a practical guide rather than a universal optimum.
Full sweeps across all adapter types and both backbones are provided in Appendix~\ref{app:full_boundary_sweep}.

%\noindent\textbf{Sensitivity-derived boundary.}
%\noindent\textbf{Comparison to random placement.} 
\noindent\textbf{Placement, not capacity, drives the gains.}
We test whether \textsc{Strap}'s gains simply reflect added capacity
(Table~\ref{tab:extra_ablations}). \textsc{Strap} outperforms both spatial adapters
in every layer with temporal adapters in the last six (12S+6T) and both types in
every layer (12S/12T), despite the latter nearly doubling the parameters, so its
gains come from placement. 12S/12T also collapses Frame2Freq whereas 12S+6T does
not, suggesting that placing both types together in early layers causes instability.
Interleaving the two types around the boundary to simulate a softer boundary (7S/7T) also underperforms.
Finally, alternative boundary rules fall short: on CLIP, the midpoint and a fixed
0.5 threshold both yield 6S/6T, trailing \textsc{Strap} by 0.6-2.9 points; on
DINOv2, the midpoint matches the \textsc{Strap} boundary, while the threshold
yields 7S/5T, which also trails it.
\vspace{-0.5em}
\subsection{Generalization to Other Backbones and Model Sizes}
\label{sec:backbone_generalization}

\noindent\begin{minipage}[t]{0.48\textwidth}
    \vspace{-5pt}
    We also test whether the $\text{\textsc{Strap}}^{\text{opt}}$ boundary is applicable beyond the DINOv2-B and CLIP-B backbones used in the main experiments.
    Table~\ref{tab:generalization} considers two extensions: larger model scales, using DINOv2-L and CLIP-L, and a different vision-only image-pretrained backbone, MAE-B~\cite{mae}.
    Across these settings, our method consistently outperforms the standard uniform adapter placement, with gains ranging from $+0.4\%$  on IKEA-ASM with DINOv2-L to $+14.5\%$ with MAE-B.
    At the same time, it reduces trainable parameters by 42--55\%.
    This suggests that the vision-only boundary is not specific to the initial models.  To further understand the generalization capability of \textsc{Strap} to other video understanding tasks, we perform experiments on temporal action detection (TAD) (Appendix~\ref{app:tad}).  On THUMOS14 with VideoMAE, \textsc{Strap} also improves over uniform placement with 22\% fewer trainable parameters.
\end{minipage}
\hfill
\begin{minipage}[t]{0.48\textwidth}
    \vspace{-5pt}
    \centering
    \captionof{table}{
    \textbf{Generalization of \textsc{Strap}.}
    \textit{Model size}: boundaries shift deeper with scale but remain identifiable and effective.
    \textit{Backbone}: VideoMAE-derived $\text{\textsc{Strap}}^{\text{opt}}=6$ transfers well to MAE-B. Results reported with ST-Adapter.
    }
    \label{tab:generalization}
    \scriptsize
    \renewcommand{\arraystretch}{0.92}
    \setlength{\tabcolsep}{3pt}
    \begin{tabular}{llccc}
        \toprule
        \textbf{Configuration} & \textbf{Strategy}
          & \textbf{Params}
          & \textbf{IKEA}
          & \textbf{HRI} \\
        \midrule
        \multicolumn{5}{l}{\textbf{\textit{Generalization to Model Size}}} \\
        \midrule
        \multicolumn{5}{l}{\textit{\textbf{DINOv2}}} \\
        
\cellcolor{gray!10}          &\cellcolor{gray!10} Uniform        & \cellcolor{gray!10}7.3  & \cellcolor{gray!10}70.5 & \cellcolor{gray!10}85.5 \\
        \cellcolor{gray!10} \multirow{-2}{*}{~~ViT-B ($\text{\textsc{Strap}}^{\text{opt}}{=}6$)}  & \cellcolor{gray!10}\textsc{Strap}& \cellcolor{gray!10}\textbf{4.2}  & \cellcolor{gray!10}\textbf{77.6} & \cellcolor{gray!10}\textbf{89.2} \\
        \addlinespace
        \multirow{2}{*}{~~ViT-L ($\text{\textsc{Strap}}^{\text{opt}}{=}16$)}
          & Uniform        & 19.0 & 80.1 & 90.3 \\
          & \textsc{Strap} & \textbf{8.5}  & \textbf{80.5} & \textbf{91.8} \\
        \midrule
        \multicolumn{5}{l}{\textit{\textbf{CLIP}}} \\
        \cellcolor{gray!10}
          & \cellcolor{gray!10}Uniform        & \cellcolor{gray!10}7.3  & \cellcolor{gray!10}70.2 & \cellcolor{gray!10}81.1 \\
        \cellcolor{gray!10}\multirow{-2}{*}{~~ViT-B ($\text{\textsc{Strap}}^{\text{opt}}{=}4$)}  & \cellcolor{gray!10}\textsc{Strap} & \cellcolor{gray!10}\textbf{5.2}  & \cellcolor{gray!10}\textbf{73.3} & \cellcolor{gray!10}\textbf{86.2} \\
        \addlinespace
        \multirow{2}{*}{~~ViT-L ($\text{\textsc{Strap}}^{\text{opt}}{=}13$)}
          & Uniform        & 19.0 & 75.6 & 88.2 \\
          & \textsc{Strap} & \textbf{10.4} & \textbf{76.7} & \textbf{89.4} \\
        \midrule
        \multicolumn{5}{l}{\textbf{\textit{Generalization to Backbone}}} \\
        \midrule
       \cellcolor{gray!10} 
          & \cellcolor{gray!10}Uniform        & \cellcolor{gray!10}7.3  & \cellcolor{gray!10}43.2 & \cellcolor{gray!10}70.5 \\
         \cellcolor{gray!10}\multirow{-2}{*}{MAE-B ($\text{\textsc{Strap}}^{\text{opt}}{=}6$)} & \cellcolor{gray!10}\textsc{Strap} & \cellcolor{gray!10}\textbf{4.2}  & \cellcolor{gray!10}\textbf{57.7} & \cellcolor{gray!10}\textbf{74.5} \\
        \bottomrule
    \end{tabular}
\end{minipage}
\par\vspace{2pt}
\flushbottom

\subsection{Generalization to Standard Action Recognition}
\label{sec:generalization}

% \subsubsection{Standard Action Recognition}
% \label{sec:coarse_action_recognition}

\noindent\textbf{Datasets.}
Beyond fine-grained action recognition, we ask whether \textsc{Strap} remains useful when the temporal signal is less central, and we evaluate coarse-grained video classification on Kinetics-400~\cite{kay2017kinetics400} and SSv2~\cite{ssv2}.
\begin{wraptable}{r}{0.5\textwidth}
   \vspace{-0.3cm}
\centering
\caption{
    \textbf{Standard (non-fine-grained) video classification results.} \textsc{Strap} reduces trainable parameters while slightly improving accuracy across most settings.
}
\label{tab:standard}
\resizebox{\linewidth}{!}{
\setlength{\tabcolsep}{2pt}
\renewcommand{\arraystretch}{1.0}
\begin{tabular}{lccccc}
\toprule
& &
\multicolumn{2}{c}{\textbf{SSv2}} &
\multicolumn{2}{c}{\textbf{K400}} \\
\cmidrule(lr){3-4} \cmidrule(lr){5-6}
\textbf{Method} &
\textbf{Strategy} &
\textbf{Params (M)} &
\textbf{Top-1 (\%)} &
\textbf{Params (M)} &
\textbf{Top-1 (\%)} \\
\midrule
\multicolumn{6}{l}{\textit{\textbf{\textit{Fully Fine-tuned Frameworks}}}} \\
\midrule
\textcolor{gray}{Video SWIN} \cite{liu2022video_swin} & \textcolor{gray}{-}
& \textcolor{gray}{88.1}
& \textcolor{gray}{69.6}
& \textcolor{gray}{88.1}
& \textcolor{gray}{80.6} \\
\textcolor{gray}{ViT-B (finetuned)} & \textcolor{gray}{-}
& \textcolor{gray}{86.1}
& \textcolor{gray}{69.3}
& \textcolor{gray}{86.1}
& \textcolor{gray}{81.0} \\
\textcolor{gray}{ILA} \cite{ILA} & \textcolor{gray}{-}
& \textcolor{gray}{-}
& \textcolor{gray}{66.8}
& \textcolor{gray}{-}
& \textcolor{gray}{81.6} \\

\textcolor{gray}{STAN-self} \cite{STAN} & \textcolor{gray}{-}
& \textcolor{gray}{-}
& \textcolor{gray}{69.5}
& \textcolor{gray}{-}
& \textcolor{gray}{84.2} \\
\midrule
\multicolumn{6}{l}{\textit{\textbf{\textit{DINOv2-B/14 backbone (6S/6T, derived from VideoMAE)}}}} \\
\midrule
\cellcolor{gray!10}
& \cellcolor{gray!10}Uniform
& \cellcolor{gray!10}14.3
& \cellcolor{gray!10}71.4
& \cellcolor{gray!10}7.3
& \cellcolor{gray!10}\textbf{81.6} \\
\multirow{-2}{*}{\cellcolor{gray!10}ST-Adapter~\cite{St_adaptor}}
& \cellcolor{gray!10}\textsc{Strap}
& \cellcolor{gray!10}\textbf{7.8} 
& \cellcolor{gray!10}\textbf{71.8}
& \cellcolor{gray!10}\textbf{4.2} 
& \cellcolor{gray!10}81.5  \\
& Uniform
& 14.3
& \textbf{71.9}
& 7.3
& 81.1 \\
\multirow{-2}{*}{Frame2Freq~\cite{frame2freq}}
& \textsc{Strap}
& \textbf{7.8} 
& 71.6 
& \textbf{4.2} 
& \textbf{81.3}  \\
\cellcolor{gray!10}
& \cellcolor{gray!10}Uniform
& \cellcolor{gray!10}28.6
& \cellcolor{gray!10}71.2
& \cellcolor{gray!10}14.8
& \cellcolor{gray!10}81.8 \\
\multirow{-2}{*}{\cellcolor{gray!10}M2-CLIP~\cite{m2_clip}}
& \cellcolor{gray!10}\textsc{Strap}
& \cellcolor{gray!10}\textbf{14.9} 
& \cellcolor{gray!10}\textbf{71.6} 
& \cellcolor{gray!10}\textbf{7.7} 
& \cellcolor{gray!10}\textbf{81.9} \\
\midrule
\multicolumn{6}{l}{\textit{\textbf{CLIP-B/16 backbone (4S/8T, derived from ViCLIP)}}} \\
\midrule
\cellcolor{gray!10}
& \cellcolor{gray!10}Uniform
& \cellcolor{gray!10}14.3
& \cellcolor{gray!10}69.5
& \cellcolor{gray!10}7.3
& \cellcolor{gray!10}82.0 \\
\multirow{-2}{*}{\cellcolor{gray!10}ST-Adapter~\cite{St_adaptor}}
& \cellcolor{gray!10}\textsc{Strap}
& \cellcolor{gray!10}\textbf{10.4} 
& \cellcolor{gray!10}\textbf{70.1}
& \cellcolor{gray!10}\textbf{5.2} 
& \cellcolor{gray!10}\textbf{82.2}  \\
& Uniform
& 14.3
& \textbf{70.4}
& 7.3
& 82.1 \\
\multirow{-2}{*}{Frame2Freq~\cite{frame2freq}}
& \textsc{Strap}
& \textbf{10.4} 
& 70.2 
& \textbf{5.2} 
& \textbf{82.2} \\
\cellcolor{gray!10}
& \cellcolor{gray!10}Uniform
& \cellcolor{gray!10}28.6
& \cellcolor{gray!10}69.1
& \cellcolor{gray!10}14.8
& \cellcolor{gray!10}81.5 \\
\multirow{-2}{*}{\cellcolor{gray!10}M2-CLIP~\cite{m2_clip}}
& \cellcolor{gray!10}\textsc{Strap}
& \cellcolor{gray!10}\textbf{19.9} 
& \cellcolor{gray!10}\textbf{70.0} 
& \cellcolor{gray!10}\textbf{9.9} 
& \cellcolor{gray!10}\textbf{82.3} \\
\bottomrule
\end{tabular}
}
\vspace{-20pt}
\end{wraptable}
Kinetics-400~\cite{kay2017kinetics400} contains ${\sim}$240k YouTube clips and 400 broad categories.
SSv2~\cite{ssv2} contains ${\sim}$220k crowd-sourced videos of predefined everyday hand-object interactions, where temporal reasoning is needed to distinguish visually similar action templates, such as pushing something left vs.\ right.  Together, they cover temporally dependent and more appearance-dominated regimes of video understanding.

\noindent\textbf{Results.}
Table~\ref{tab:standard} shows that across adapters and backbones, STRAP reduces the number of trainable parameters substantially, by up to 48\%, while matching or slightly improving accuracy in most settings.
The improvements are naturally smaller than on fine-grained benchmarks, where temporal adaptation is more critical.
Another possible explanation is that datasets such as Kinetics-400 contain many web videos in which object, scene, and appearance cues are already well covered by image-pretrained representations, making extensive video-specific adaptation less critical than in fine-grained settings that depend more strongly on subtle temporal structure.
However, the consistent parameter reduction with stable accuracy suggests that optimizing adapter placement is a simple and robust strategy to at least boost the efficiency of the mechanism in standard video classification.

\section{Discussion}
\label{sec:experiment_discussion}

% \noindent\textbf{If video-pretrained models guide adapter placement in image-to-video transfer, why not use them for recognition directly?}
% Since \textsc{Strap} uses video-pretrained models as guides for adapter placements in image-to-video PEFT, a natural question arises: why adapt image-pretrained models at all, rather than using video-pretrained models directly?
% We see two main reasons.
% First, image-to-video PEFT is substantially more parameter-efficient.
% Video-pretrained models operate natively on spatio-temporal inputs and typically require heavier architectures or full-model adaptation, whereas image-to-video PEFT keeps a strong image backbone frozen and only trains minimal adapters.\red{Maybe add an example here, how many parameters one image vs. video model has.}
% Second, image-based backbones, despite being smaller, often deliver better recognition results once equipped with suitable temporal modules.
% In Table~\ref{tab:finegrained}, \textsc{Strap} reaches state-of-the-art results on all four fine-grained benchmarks among the compared methods, including fully fine-tuned video models shown for context.
% This is also consistent with prior image-to-video PEFT work \red{Cite if true}, where adapted image models can match or outperform fully video-pretrained alternatives. 
% Video models, such as VideoSwin,  are a natural fit for temporal structure, but image foundation models often benefit from larger and more diverse pretraining corpora.

\noindent\textbf{Why should the temporal sensitivity of video-pretrained models transfer to image-to-video PEFT?}
We do not assume that representations transfer from video to image models. Instead, \textsc{Strap} uses a weaker structural prior derived from our analysis in Section~\ref{sec:analysis}: temporal sensitivity emerges non-uniformly across depth, and its emergence profile depends on the pretraining paradigm. 
% This is motivated by the sensitivity analysis in Section~\ref{sec:analysis}, where vision-only and vision--language video models show different depth-wise temporal sensitivity profiles.
% Rather than transferring specific features from the video-pretrained reference model, \textsc{Strap} transfers only its depth-wise temporal structure.
Concretely, we derive the boundary ${\textsc{Strap}}^{\text{opt}}$ from the 
temporal sensitivity profile of a frozen, paradigm-matched video model:  a vision-only proxy for DINOv2 and a vision-language proxy for CLIP. 
% evolution of the temporal sensitivity ratio $R(l)$ across layers.
% By matching the reference model to the target backbone paradigm, vision-only for DINOv2 and vision-language for CLIP, we use this structural transition as a placement prior for image-to-video PEFT.
The empirical results support this: the derived boundaries perform well in the allocation sweep (Fig.~\ref{fig:ablation}), improve over uniform placement across adapters and datasets (Table~\ref{tab:finegrained}), and hold across additional backbones and model scales (Table~\ref{tab:generalization}).
This suggests that \textsc{Strap} is not simply a balanced-split heuristic: the boundary is mid-depth for DINOv2-B, earlier for CLIP-B, and shifts with scale (Table~\ref{tab:generalization}), reflecting paradigm- and scale-specific sensitivity profiles without validation data.

\noindent\textbf{Is the boundary a principled choice or a tuned hyperparameter?}
The boundary is not tuned on downstream validation data.
It is derived from the peak change in the temporal sensitivity ratio of a paradigm-matched video-pretrained reference model, as defined in Section~\ref{sec:analysis}.
The allocation sweep in Fig.~\ref{fig:ablation} shows that these sensitivity-derived boundaries are best or near-best among spatial-temporal splits for both DINOv2 and CLIP. We therefore view ${\textsc{Strap}}^{\text{opt}}$ as a training-free placement rule, while acknowledging that the exact best split can vary slightly across datasets.

\noindent\textbf{When does \textsc{Strap} help for recognition quality, and when is it mainly an efficiency gain?}
\textsc{Strap} yields the largest accuracy gains on fine-grained video tasks (Table~\ref{tab:finegrained}), where subtle temporal differences are often class-discriminative. This is not inherited from existing image-to-video adapters: prior adapters (with the exception of Frame2Freq \cite{frame2freq}) struggle to consistently improve fine-grained activity recognition \cite{frame2freq}. \textsc{Strap} improves this setting by changing \emph{where} the adapters are placed, rather than introducing a new temporal adapter. 
% This is consistent with prior image-to-video PEFT work, where temporal adapters tend to be most beneficial for fine-grained video classification rather than broader coarse-grained recognition.\red{Todo: Cite if true, otherwise remove.}
%One explanation is that fine-grained classes often share objects, scenes, and appearance, and therefore require subtle motion cues, temporal ordering, or interaction dynamics.
The symmetric-action analysis in Table~\ref{tab:sym_nonsym_wrap} supports this: improvements are largest for visually similar classes that differ mainly in temporal direction, suggesting that existing adapters are limited not only by their design but also by where temporal capacity is placed.
On standard benchmarks such as Kinetics-400~\cite{kay2017kinetics400}, where appearance cues are often sufficient, \textsc{Strap} mainly acts as an efficiency mechanism, reducing trainable parameters without sacrificing accuracy (Table~\ref{tab:standard}).
% mainly acts as an efficiency mechanism. This is primarily due to the dominance of spatial cues in these datasets
% the gains are smaller, and \textsc{Strap} mainly acts as an efficiency mechanism (Table~\ref{tab:standard}) 
% - a likely consequence of the appearance-rich web videos in datasets such as Kinetics-400 \cite{kay2017kinetics400}. 
% On standard action recognition benchmarks, the gains are smaller, and \textsc{Strap} mainly acts as an efficiency mechanism (Table~\ref{tab:standard}).
% A likely reason is that datasets such as Kinetics-400 \cite{kay2017kinetics400} are collected from diverse web videos, where large-scale image pretraining already provides strong appearance and scene representations.
% In such settings, extensive temporal adaptation may be less critical for accuracy, although better placement still reduces trainable parameters without sacrificing performance.
Thus, \textsc{Strap} is most beneficial for accuracy when temporal structure is discriminative, and mainly improves efficiency when appearance cues dominate.
%In few-shot recognition, \textsc{Strap} improves all 18 evaluated settings (Table~\ref{tab:fewshot}), suggesting that better placement is also useful when labeled data are scarce.

%% file: table_main_fine_grained.tex
\newcommand{\dpos}[1]{\textcolor{green!60!black}{\tiny(\textbf{+#1})}}
\newcommand{\dneg}[1]{\textcolor{red!70!black}{\tiny(#1)}}
\newcommand{\dpospar}[1]{\textcolor{green!60!black}{\scriptsize(\textbf{$\downarrow$#1})}}

\begin{table*}[t!]
\centering
\scriptsize
\caption{\textbf{Fine-grained action recognition.}
\textsc{STRAP} improves accuracy (up to +11.8\%) while reducing trainable parameters by up to 48\% and inference time by up to 10\%. Gains are largest on temporally demanding datasets (D\&A, IKEA, HRI) and consistent across methods and backbones.}
\label{tab:finegrained}
\renewcommand{\arraystretch}{1.0}
\setlength{\tabcolsep}{4pt}
\begin{tabular}{lccc|cccc}
\toprule
& &
\multirow{2}{*}{\textbf{Params}} &
\multirow{2}{*}{\textbf{Inf.\ Time}} &
\multicolumn{4}{c}{\textbf{Top-1 Accuracy (\%) $\uparrow$}} \\
\cmidrule(lr){5-8}
\textbf{Method} &
\textbf{Strategy} &
\textbf{(M)} & \textbf{(ms)$^\dagger$} &
\textbf{Diving} &
\textbf{Drive\&Act} &
\textbf{IKEA} &
\textbf{HRI-30} \\
\midrule
\multicolumn{8}{l}{\textit{\textbf{Fully Fine-tuned Frameworks}}} \\
\midrule
\textcolor{gray}{ORViT} \cite{OR_vit}
& \textcolor{gray}{-}
& \textcolor{gray}{160}
& \textcolor{gray}{-}
& \textcolor{gray}{88.0}
& \textcolor{gray}{-}
& \textcolor{gray}{-}
& \textcolor{gray}{-} \\

\textcolor{gray}{VideoSwin-B } \cite{liu2022video_swin}
& \textcolor{gray}{-}
& \textcolor{gray}{88}
& \textcolor{gray}{-}
& \textcolor{gray}{-}
& \textcolor{gray}{72.9}
& \textcolor{gray}{72.6}
& \textcolor{gray}{-} \\

\textcolor{gray}{UniformerV2} \cite{uniformerv2}
& \textcolor{gray}{-}
& \textcolor{gray}{115}
& \textcolor{gray}{-}
& \textcolor{gray}{-}
& \textcolor{gray}{76.7}
& \textcolor{gray}{-}
& \textcolor{gray}{-} \\

\textcolor{gray}{SlowOnly} \cite{hri_30}
& \textcolor{gray}{-}
& \textcolor{gray}{32}
& \textcolor{gray}{-}
& \textcolor{gray}{-}
& \textcolor{gray}{-}
& \textcolor{gray}{-}
& \textcolor{gray}{86.6} \\
\midrule
\multicolumn{8}{l}{\textit{\textbf{DINOv2-B/14 backbone (6S/6T, derived from VideoMAE)}}} \\
\midrule

\cellcolor{gray!10}
& \cellcolor{gray!10}Uniform
& \cellcolor{gray!10}7.3
& \cellcolor{gray!10}54.5
& \cellcolor{gray!10}88.6
& \cellcolor{gray!10}75.2
& \cellcolor{gray!10}70.5
& \cellcolor{gray!10}85.5 \\
\multirow{-2}{*}{\cellcolor{gray!10}ST-Adapter~\cite{St_adaptor}}
& \cellcolor{gray!10}\textsc{Strap}
& \cellcolor{gray!10}\textbf{4.2} \dpospar{42\%}
& \cellcolor{gray!10}\textbf{50.8} \dpospar{6.8\%}
& \cellcolor{gray!10}\textbf{88.9} \dpos{0.3}
& \cellcolor{gray!10}\textbf{81.9} \dpos{6.7}
& \cellcolor{gray!10}\textbf{77.6} \dpos{7.1}
& \cellcolor{gray!10}\textbf{89.2} \dpos{3.7} \\

& Uniform
& 7.3
& 57.7
& 91.7
& 81.5
& 76.4
& 89.8 \\
\multirow{-2}{*}{Frame2Freq~\cite{frame2freq}}
& \textsc{Strap}
& \textbf{4.2} \dpospar{42\%}
& \textbf{51.7} \dpospar{10.4\%}
& \textbf{92.5} \dpos{0.8}
& \textbf{83.5} \dpos{2.0}
& \textbf{79.8} \dpos{3.4}
& \textbf{91.2} \dpos{1.4} \\

\cellcolor{gray!10}
& \cellcolor{gray!10}Uniform
& \cellcolor{gray!10}14.8
& \cellcolor{gray!10}64.6
& \cellcolor{gray!10}90.0
& \cellcolor{gray!10}77.7
& \cellcolor{gray!10}74.9
& \cellcolor{gray!10}80.5 \\
\multirow{-2}{*}{\cellcolor{gray!10}M2-CLIP~\cite{m2_clip}}
& \cellcolor{gray!10}\textsc{Strap}
& \cellcolor{gray!10}\textbf{7.7} \dpospar{48\%}
& \cellcolor{gray!10}\textbf{57.8} \dpospar{10.4\%}
& \cellcolor{gray!10}\textbf{90.3} \dpos{0.3}
& \cellcolor{gray!10}\textbf{82.9} \dpos{5.2}
& \cellcolor{gray!10}\textbf{79.6} \dpos{4.7}
& \cellcolor{gray!10}\textbf{92.3} \dpos{11.8} \\

\midrule
\multicolumn{8}{l}{\textit{\textbf{CLIP-B/16 backbone (4S/8T, derived from ViCLIP)}}} \\
\midrule

\cellcolor{gray!10}
& \cellcolor{gray!10}Uniform
& \cellcolor{gray!10}7.3
& \cellcolor{gray!10}46.7
& \cellcolor{gray!10}90.4
& \cellcolor{gray!10}72.6
& \cellcolor{gray!10}70.2
& \cellcolor{gray!10}81.1 \\
\multirow{-2}{*}{\cellcolor{gray!10}ST-Adapter~\cite{St_adaptor}}
& \cellcolor{gray!10}\textsc{Strap}
& \cellcolor{gray!10}\textbf{5.2} \dpospar{29\%}
& \cellcolor{gray!10}\textbf{44.2} \dpospar{5.5\%}
& \cellcolor{gray!10}\textbf{90.5} \dpos{0.1}
& \cellcolor{gray!10}\textbf{80.8} \dpos{8.2}
& \cellcolor{gray!10}\textbf{73.3} \dpos{3.1}
& \cellcolor{gray!10}\textbf{86.2} \dpos{5.1} \\

& Uniform
& 7.3
& 49.6
& \textbf{92.2}
& 77.4
& \textbf{78.0}
& 85.7 \\
\multirow{-2}{*}{Frame2Freq~\cite{frame2freq}}
& \textsc{Strap}
& \textbf{5.2} \dpospar{29\%}
& \textbf{47.2} \dpospar{4.9\%}
& 91.1 \dneg{-1.1}
& \textbf{81.5} \dpos{4.1}
& 76.5 \dneg{-1.5}
& \textbf{86.3} \dpos{0.6} \\

\cellcolor{gray!10}
& \cellcolor{gray!10}Uniform
& \cellcolor{gray!10}14.8
& \cellcolor{gray!10}52.3
& \cellcolor{gray!10}89.3
& \cellcolor{gray!10}77.2
& \cellcolor{gray!10}71.9
& \cellcolor{gray!10}85.9 \\
\multirow{-2}{*}{\cellcolor{gray!10}M2-CLIP~\cite{m2_clip}}
& \cellcolor{gray!10}\textsc{Strap}
& \cellcolor{gray!10}\textbf{9.9} \dpospar{33\%}
& \cellcolor{gray!10}\textbf{48.8} \dpospar{6.7\%}
& \cellcolor{gray!10}\textbf{90.3} \dpos{1.0}
& \cellcolor{gray!10}\textbf{82.9} \dpos{5.7}
& \cellcolor{gray!10}\textbf{76.2} \dpos{4.3}
& \cellcolor{gray!10}\textbf{93.1} \dpos{7.2} \\

\bottomrule
\end{tabular}
 \vspace{4pt}

\noindent{\footnotesize \textbf{$^{\dagger}$}Inference time per sample (ms) measured on a single NVIDIA A100 GPU, averaged over 100 runs.}
\vspace{-1em}
\end{table*}

%% file: table_symmetric.tex
\begin{wraptable}{r}{0.5\textwidth}
\centering
\vspace{-15pt} % Adjust to align with text
\caption{
    \textbf{Results for symmetric and non-symmetric actions.}
    \textit{Sym} and \textit{N-S} denote accuracy on symmetric and non-symmetric action subsets. Our method has the largest gains for symmetric actions (e.g., \textit{opening} vs. \textit{closing}), which require distinguishing temporal order and direction, precisely the structure targeted by late-layer temporal adapters.
}
\label{tab:sym_nonsym_wrap}
\resizebox{\linewidth}{!}{
\setlength{\tabcolsep}{2pt}
\renewcommand{\arraystretch}{1.1}
\begin{tabular}{lc ccc | ccc | ccc}
\toprule
& & \multicolumn{3}{c}{\textbf{Drive\&Act}} & \multicolumn{3}{c}{\textbf{IKEA-ASM}} & \multicolumn{3}{c}{\textbf{HRI-30}} \\
\cmidrule(lr){3-5} \cmidrule(lr){6-8} \cmidrule(lr){9-11}
\textbf{Method} & \textbf{Strat.} & Sym & N-S & All & Sym & N-S & All & Sym & N-S & All \\
\midrule
\cellcolor{gray!10}\multirow{2}{*}{ST-Adapter} & \cellcolor{gray!10}Uni. & \cellcolor{gray!10}66.4 & \cellcolor{gray!10}83.4 & \cellcolor{gray!10}75.2 & \cellcolor{gray!10}68.5 & \cellcolor{gray!10}73.9 & \cellcolor{gray!10}70.5 & \cellcolor{gray!10}81.5 & \cellcolor{gray!10}95.3 & \cellcolor{gray!10}85.5 \\
                \cellcolor{gray!10}\multirow{-2}{*}{ST-Adapter}             & \cellcolor{gray!10}\textsc{Strap} & \cellcolor{gray!10}\textbf{78.0} & \cellcolor{gray!10}\textbf{85.5} & \cellcolor{gray!10}\textbf{81.9} & \cellcolor{gray!10}\textbf{80.3} & \cellcolor{gray!10}\textbf{76.2} & \cellcolor{gray!10}\textbf{77.6} & \cellcolor{gray!10}\textbf{85.4} & \cellcolor{gray!10}\textbf{96.8} & \textbf{89.2} \\
\addlinespace
\multirow{2}{*}{Frame2Freq} & Uni. & 77.1 & 85.5 & 81.5 & 81.0 & 74.2 & 76.4 & 85.6 & \textbf{98.9} & 89.8 \\
                             & \textsc{Strap} & \textbf{79.6} & \textbf{87.1} & \textbf{83.5} & \textbf{86.8} & \textbf{76.4} & \textbf{79.8} & \textbf{88.2} & 97.2 & \textbf{91.2} \\
\addlinespace
\cellcolor{gray!10}  & \cellcolor{gray!10}Uni. & \cellcolor{gray!10}68.4 & \cellcolor{gray!10}85.0 & \cellcolor{gray!10}77.7 & \cellcolor{gray!10}75.3 & \cellcolor{gray!10}74.6 & \cellcolor{gray!10}74.9 & \cellcolor{gray!10}78.4 & \cellcolor{gray!10}84.6 & \cellcolor{gray!10}80.5 \\
                           \cellcolor{gray!10}\multirow{-2}{*}{M2-CLIP}    & \cellcolor{gray!10}\textsc{Strap} & \cellcolor{gray!10}\textbf{78.8} & \cellcolor{gray!10}\textbf{86.6} & \cellcolor{gray!10}\textbf{82.9} & \cellcolor{gray!10}\textbf{83.7} & \cellcolor{gray!10}\textbf{77.6} & \cellcolor{gray!10}\textbf{79.6} & \cellcolor{gray!10}\textbf{89.3} & \cellcolor{gray!10}\textbf{98.2} & \cellcolor{gray!10}\textbf{92.3} \\
\bottomrule
\end{tabular}
}
\vspace{-15pt}
\end{wraptable}

%% file: ablation_tables.tex
\begin{figure}[t]
\begin{minipage}[t]{0.48\textwidth}
    \vspace{0pt}
    \centering
    \captionof{table}{
    \textbf{Adapter placement ablation.} We isolate: 
    (\textit{i})~component contribution - does each adapter type help? 
    (\textit{ii})~placement direction - does spatial-before-temporal matter? 
    (\textit{iii})~capacity control - is the gain from placement or parameter count?
    Results reported using Frame2Freq Adapter and 
    DINOv2 ($\text{\textsc{Strap}}^{\text{opt}}=6$).
    %$\star$ marks the \textsc{Strap}-predicted configuration.
    }
    \label{tab:component_ablation}
    \scriptsize
    \renewcommand{\arraystretch}{1.0}
    \setlength{\tabcolsep}{3pt}
    \begin{tabular}{lcccc}
        \toprule
        \textbf{Configuration}
          & \textbf{Params}
          & \textbf{D\&A}
          & \textbf{IKEA}
          & \textbf{HRI} \\
          
        \midrule
                \multicolumn{5}{l}{\textbf{\textit{Placement direction}}} \\
        \midrule
          \cellcolor{gray!10}Reversed \scriptsize{(T$\rightarrow$S)}
            & \cellcolor{gray!10}4.2 & \cellcolor{gray!10}72.4 &\cellcolor{gray!10} 39.1 & \cellcolor{gray!10}75.5 \\
          \textbf{\textsc{Strap}} \scriptsize{(S$\rightarrow$T)} 
            & \textbf{4.2}
            & \textbf{83.5}
            & \textbf{79.8}
            & \textbf{91.2} \\
        \midrule
        
        \multicolumn{5}{l}{\textbf{\textit{Component}}} \\
        \midrule
          \cellcolor{gray!10}Spatial only \tiny{(layers 1-6)}
            & \cellcolor{gray!10}0.6 & \cellcolor{gray!10}71.7 & \cellcolor{gray!10}66.6 &\cellcolor{gray!10} 51.1 \\
          Temporal only \tiny{(layers 7-12)}
            & 3.6 & 81.5 & 78.0 & 90.1 \\
          \cellcolor{gray!10}\textbf{\textsc{Strap}} \tiny{(Ours, 6S+6T)} 
            & \cellcolor{gray!10}\textbf{4.2}
            & \cellcolor{gray!10}\textbf{83.5}
            & \cellcolor{gray!10}\textbf{79.8}
            & \cellcolor{gray!10}\textbf{91.2} \\
        \midrule

        \multicolumn{5}{l}{\textbf{\textit{Capacity control (matched budget)}}}\\
        \midrule
         \cellcolor{gray!10}Uniform Temporal
            &\cellcolor{gray!10} 7.3 & \cellcolor{gray!10}81.5 & \cellcolor{gray!10}76.4 & \cellcolor{gray!10}89.8 \\
          Uniform Temporal \tiny{(adjusted width)}
            & 4.2 & 70.2 & 76.1 & 87.7 \\
          \cellcolor{gray!10}\textbf{\textsc{Strap}} \scriptsize{(Ours)} 
            & \cellcolor{gray!10}\textbf{4.2}
            & \cellcolor{gray!10}\textbf{83.5}
            & \cellcolor{gray!10}\textbf{79.8}
            & \cellcolor{gray!10}\textbf{91.2} \\
        \bottomrule
    \end{tabular}
\end{minipage}
\hfill
\begin{minipage}[t]{0.48\textwidth}
    \vspace{0pt}
    \centering
  
    \captionof{table}{\textbf{Placement, interleaving and boundary-rule ablations.}
  Accuracy (\%) on IKEA-ASM and HRI-30.
  \textsc{Strap} performs best in every column, with the fewest parameters on DINOv2.}
    \label{tab:extra_ablations}
    % TODO: red Params are derived from the per-layer costs in tab:component_ablation
    % (about 0.1M per spatial layer, 0.6M per temporal layer). Replace with measured
    % counts and remove .
    \scriptsize
    \renewcommand{\arraystretch}{1.0}
    \setlength{\tabcolsep}{3pt}
    \begin{tabular}{lccccc}
    \toprule
     & & \multicolumn{2}{c}{\textbf{Frame2Freq}} & \multicolumn{2}{c}{\textbf{M2-CLIP}} \\
    \cmidrule(lr){3-4}\cmidrule(lr){5-6}
    \textbf{Configuration} & \textbf{Params} & \textbf{IKEA} & \textbf{HRI} & \textbf{IKEA} & \textbf{HRI} \\
    \midrule
    \multicolumn{6}{l}{\textbf{\textit{Placement vs.\ capacity (DINOv2)}}} \\
    \midrule
    \cellcolor{gray!10}Uniform \tiny{(12T)}
      & \cellcolor{gray!10}7.3 & \cellcolor{gray!10}76.4 & \cellcolor{gray!10}89.8 & \cellcolor{gray!10}74.9 & \cellcolor{gray!10}80.5 \\
    12S+6T
      & {4.8} & 78.1 & 88.2 & 78.4 & 90.3 \\
    \cellcolor{gray!10}12S/12T
      & \cellcolor{gray!10}{8.5} & \cellcolor{gray!10}41.1 & \cellcolor{gray!10}72.9 & \cellcolor{gray!10}78.6 & \cellcolor{gray!10}91.2 \\
    \textbf{\textsc{Strap}} \tiny{(6S/6T)}
      & \textbf{4.2} & \textbf{79.8} & \textbf{91.2} & \textbf{79.6} & \textbf{92.3} \\
    \midrule
    \multicolumn{6}{l}{\textbf{\textit{Interleaving (DINOv2)}}} \\
    \midrule
    \cellcolor{gray!10}Interleaved \tiny{(7S/7T)}
      & \cellcolor{gray!10}{4.9} & \cellcolor{gray!10}77.8 & \cellcolor{gray!10}88.4 & \cellcolor{gray!10}78.6 & \cellcolor{gray!10}90.2 \\
    \textbf{\textsc{Strap}} \tiny{(6S/6T)}
      & \textbf{4.2} & \textbf{79.8} & \textbf{91.2} & \textbf{79.6} & \textbf{92.3} \\
    \midrule
    \multicolumn{6}{l}{\textbf{\textit{Boundary rule (CLIP)}}} \\
    \midrule
    \cellcolor{gray!10}Midpoint \tiny{(6S/6T)}
      & \cellcolor{gray!10}{4.2} & \cellcolor{gray!10}74.2 & \cellcolor{gray!10}83.4 & \cellcolor{gray!10}75.6 & \cellcolor{gray!10}92.2 \\
    Threshold = 0.5 \tiny{(6S/6T)}
      & {4.2} & 74.2 & 83.4 & 75.6 & 92.2 \\
    \cellcolor{gray!10}\textbf{\textsc{Strap}} \tiny{(4S/8T)}
      & \cellcolor{gray!10}{\textbf{5.2}} & \cellcolor{gray!10}\textbf{76.5} & \cellcolor{gray!10}\textbf{86.3} & \cellcolor{gray!10}\textbf{76.2} & \cellcolor{gray!10}\textbf{93.1} \\
    \bottomrule
    \end{tabular}
\end{minipage}
\end{figure}

%% file: supplemental.tex
\newpage
\appendix
\section{Technical appendix}
This appendix provides additional implementation details and analyses supporting the main paper.
We first describe the training and evaluation setup in Section~\ref{app:implementation}, then validate the sensitivity metric through leave-one-out perturbation analysis in Section~\ref{app:perturbation_robustness}. We extend our sensitivity analysis to four more Video backbones in Section \ref{app:add_backbones}. 
Section~\ref{app:full_boundary_sweep} reports full spatial--temporal boundary sweeps across adapter families and backbones. 
Finally, Sections~\ref{app:fewshot} and~\ref{app:sym_nonsym} present few-shot results and an extended symmetric/non-symmetric action breakdown with CLIP-based models. To understand generalization to other video understanding tasks, \ref{app:tad} presents Temporal Action Detection (TAD) results.

%%%%%%%%%%%%%%%%%%%%%%%%%%%%%%%%%%%%%%%%%%%%%%%%%%%%%%%%%%%%
\subsection{Implementation Details}
\label{app:implementation}

\noindent\textbf{Backbones and adapter placement.}
Unless stated otherwise, all experiments use frozen DINOv2-B/14 and CLIP-B/16 as image-pretrained backbones.
Adapter modules are inserted into each transformer block following the placement convention of the respective base method: ST-Adapter~\cite{St_adaptor}, Frame2Freq~\cite{frame2freq}, and M2-CLIP~\cite{m2_clip}.
\textsc{Strap} modifies only the layer-wise assignment of spatial and temporal adapters.
Temporal adapters use a bottleneck dimension of 384 following the base methods, while spatial adapters use a bottleneck dimension of 64 following AdaptFormer~\cite{adaptformer}.
Only the adapter modules and a linear classifier are trained; the image-pretrained backbone remains frozen throughout.

\noindent\textbf{Optimization.}
All models are trained with AdamW for 60 epochs, using weight decay $0.001$ and label smoothing $0.1$.
We use the learning rates specified by respective adapters for CLIP and a standard learning rate of $1\times10^{-3}$ for DINOv2 backbone.
We apply linear warmup for the first 5 epochs on Diving48, Kinetics-400, and SSv2, and for the first 2 epochs on Drive\&Act, IKEA-ASM, and HRI-30. The batch size is 8 per GPU.

\noindent\textbf{Hardware.}
Training uses 4 NVIDIA A100 GPUs for Drive\&Act, IKEA-ASM, and HRI-30, and 16 NVIDIA A100 GPUs for Diving48, Kinetics-400, and SSv2.

\noindent\textbf{Data preprocessing and sampling.}
All models use inputs at $224\times224$ resolution.
We use 16 frames for Drive\&Act, IKEA-ASM, and  HRI-30,  8 frames for  Kinetics-400, and 32 frames for SSv2 and Diving48.
At inference, we use three temporal clips for Drive\&Act, IKEA-ASM, HRI-30, and Kinetics-400, and three spatial crops with uniform sampling for Diving48, Kinetics-400, and SSv2.
Official dataset splits are used throughout.

\noindent\textbf{Few-shot evaluation.}
Few-shot action recognition follows the standard $N$-way $K$-shot episodic protocol of D2-ST Adapter~\cite{D2_ST}, including the number of evaluation episodes per setting.

\noindent\textbf{Sensitivity analysis.}
The temporal sensitivity profile $R(l)$ is computed using 500 unlabeled clips per model--dataset pair, sampled from the training split. We use 6 datasets:  Kinetics \cite{kay2017kinetics400}, SSv2\cite{ssv2}, Diving48 \cite{diving48}, Drive\&Act \cite{drive_and_act}, IKEA-ASM \cite{IKEA_ASM} and HRI-30 \cite{hri_30}. 
For each proxy model, we compute $R(l)$ independently for each dataset, normalize the resulting profile per model--dataset pair, and average the profiles across all six evaluation datasets.
We then compute $\Delta R(l)$ on the averaged profile and select $\text{\textsc{Strap}}^{\text{opt}}$ as the layer with the maximum increase.
This yields a single backbone-specific boundary per proxy model, which is applied uniformly across all downstream tasks without per-dataset tuning.
The cost of deriving $R(l)$ is equivalent to standard inference over the sampled clips for the corresponding proxy model and is incurred only once; it does not affect training or evaluation.

\subsection{Perturbation Robustness Analysis}
\label{app:perturbation_robustness}
\begin{figure}[ht]
	\centering
\includegraphics[width=1\textwidth]{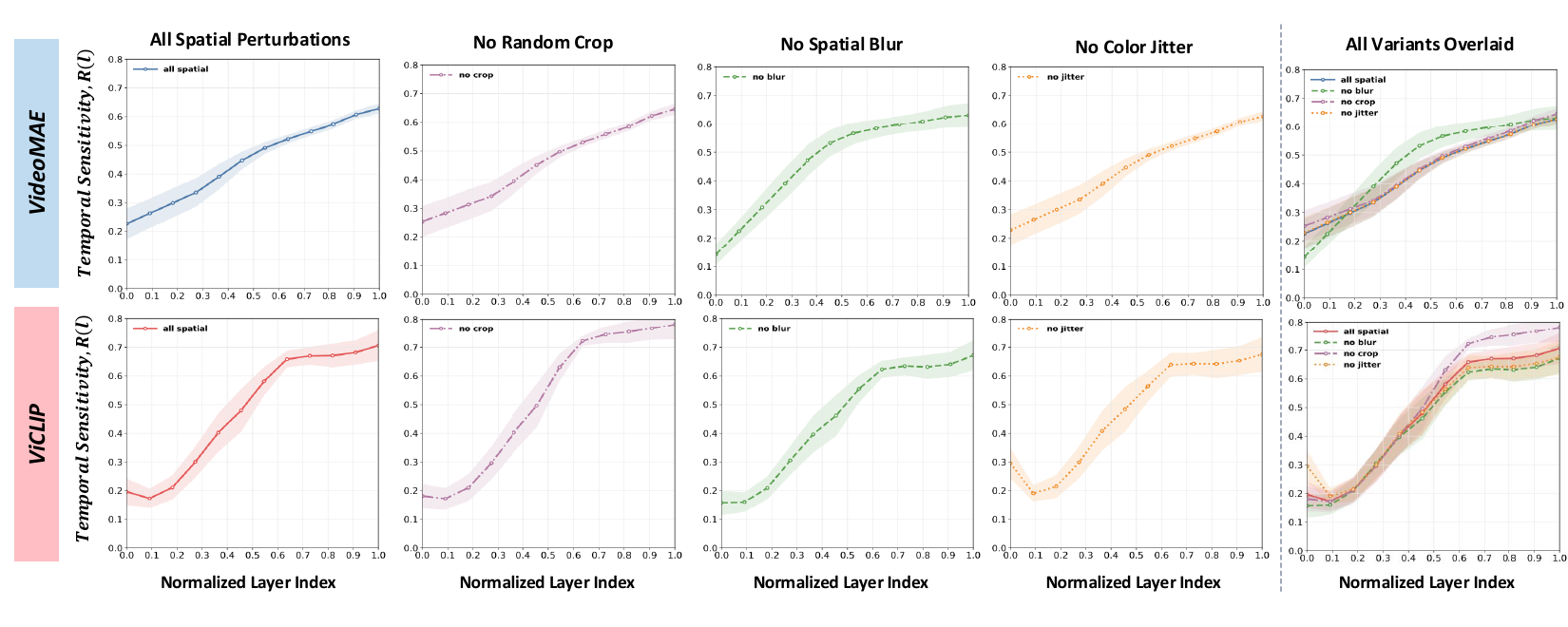}
\vspace{-1em}
    \caption{\textbf{Impact of spatial perturbation choice on
$R(l)$}. Leave-one-out analysis removing each spatial perturbation in turn: random crop, spatial blur, and color jitter. The depth-wise profile and paradigm-specific signatures remain consistent across all variants for both VideoMAE and ViCLIP.}
\vspace{-1em}
	\label{fig:spatial_robustness}

\end{figure}

\begin{figure}[ht]
	\centering
\includegraphics[width=1\textwidth]{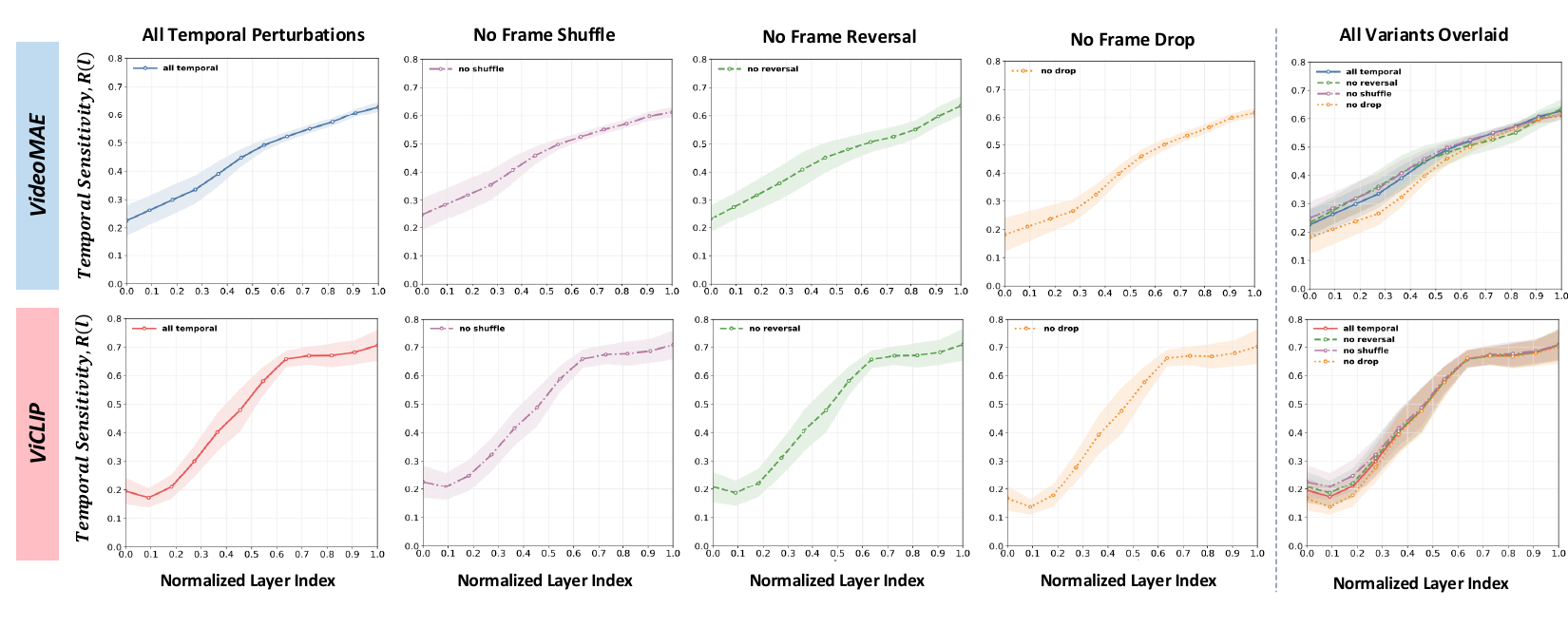}
\vspace{-1em}
    \caption{\textbf{Impact of temporal perturbation choice on $R(l)$}. Leave-one-out analysis removing each temporal perturbation in turn: frame shuffle, frame reversal, and frame drop. The depth-wise profile and paradigm-specific signatures remain stable across all variants for both VideoMAE and ViCLIP.}
\vspace{-1em}
	\label{fig:temporal_robustness}

\end{figure}

We analyze the sensitivity of $R(l)$ to the perturbation set using a leave-one-out analysis.
For spatial perturbations, we remove random crop, spatial blur, and color jitter in turn; for temporal perturbations, we remove frame shuffle, frame reversal, and frame drop in turn.
Figures~\ref{fig:spatial_robustness} and~\ref{fig:temporal_robustness} show the resulting profiles for VideoMAE and ViCLIP.

Across all variants, the depth-wise profile of $R(l)$ remains stable.
For VideoMAE, the gradual increase toward later layers is preserved under all perturbation subsets.
For ViCLIP, the characteristic early suppression followed by late recovery remains visible regardless of which perturbation is removed.
The overlaid profiles show tight clustering across leave-one-out variants, indicating that the observed spatial--temporal transition is not driven by a single perturbation choice.

\subsection{Sensitivity Analysis for Additional Video Backbones}
\label{app:add_backbones} 
\begin{figure}[ht!]
    \centering
    \includegraphics[width=\linewidth]{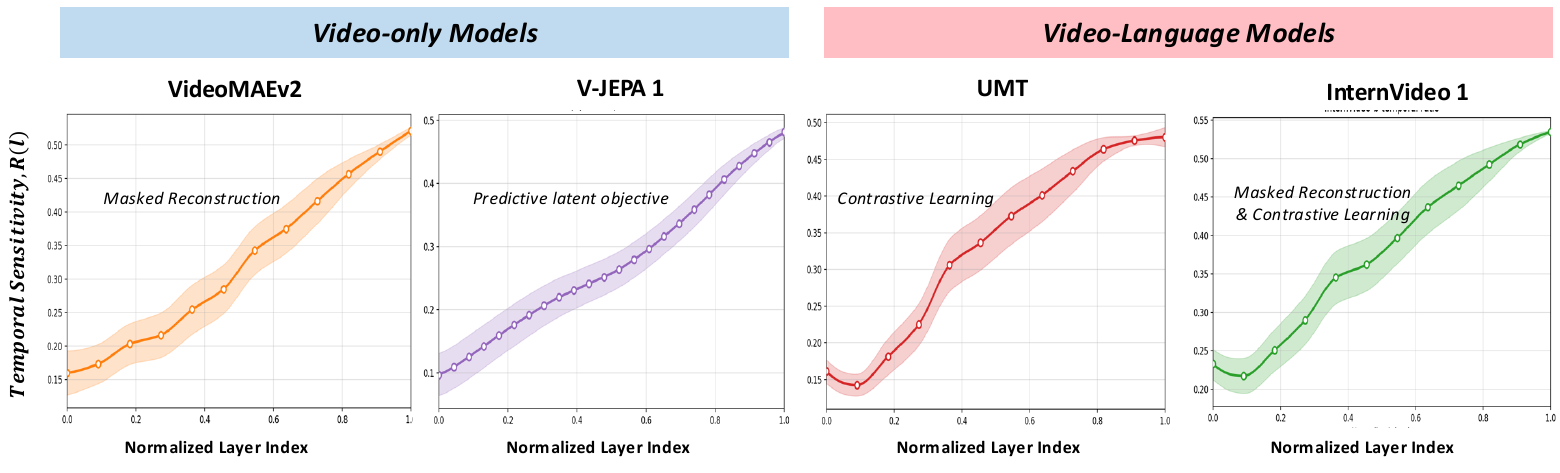}
    \caption{Layer-wise temporal sensitivity $R(l)$ of four additional video foundation models. $R(l)$ increases with depth, and its growth profile depends on the pretraining paradigm.}
    \label{fig:sensitivity_additional}
\end{figure}

To check that our findings generalize, we repeat the sensitivity analysis of Sec.~3 on four additional video foundation models: two video-only models, VideoMAEv2~\cite{videomaev2} (masked reconstruction) and V-JEPA1~\cite{v-jepa} (predictive latent), and two video-language models, UMT~\cite{UMT} (contrastive) and InternVideo~1~\cite{internvideo} (masked reconstruction and contrastive). The setup is unchanged. As Fig.~\ref{fig:sensitivity_additional} shows, both findings hold. $R(l)$ increases with depth in all four models (Finding~1). The video-only models rise gradually, whereas the video-language models show an early dip before recovering (Finding~2).

% \subsection{Sensitivity Analysis for additional Video backbones}
% \label{app:add_backbones} 

% \begin{figure}[ht]
% 	\centering
% \includegraphics[width=1\textwidth]{Supplemental_sensitvity_analysis.pdf}
% \vspace{-1em}
% \caption{Layer-wise temporal sensitivity $R(l)$ of four video foundation models.
% Depth increases across layers, and the growth profile is dependent on the pre-training paradigm.}
% \vspace{-1em}
% 	\label{fig:temporal_robustness}

% \end{figure}
\subsection{Full Boundary Sweep Results}
\label{app:full_boundary_sweep} 

To further analyze the effect of the spatial--temporal boundary, we provide full boundary sweeps for all three adapter families: ST-Adapter, Frame2Freq, and M2-CLIP.
For each method and backbone, we vary the number of spatial and temporal adapters while keeping the adapter width fixed.
This produces configurations ranging from all-spatial adaptation to all-temporal adaptation, with intermediate spatial--temporal splits such as 10S/2T, 8S/4T, 6S/6T, 4S/8T, and 2S/10T.
The \textsc{Strap}-predicted boundary is marked with $^\dagger$.

Across adapter families, the sweep confirms that adapter placement has a strong effect on performance.
All-spatial configurations are consistently weak, showing that temporal modeling is necessary for fine-grained video recognition.
At the same time, all-temporal placement is also often suboptimal, suggesting that applying temporal modules uniformly across depth can misallocate capacity.
Intermediate spatial--temporal splits generally provide the best accuracy--efficiency trade-off.

For DINOv2, the sensitivity-derived 6S/6T split is consistently competitive and often achieves the best or near-best performance across datasets and adapter types.
For CLIP, the best region shifts toward a later temporal allocation, with the 4S/8T split performing strongest in most settings.
This agrees with the sensitivity analysis: the vision-only proxy predicts a later spatial--temporal transition for DINOv2, while the vision-language proxy predicts an earlier transition for CLIP.
Importantly, using the same boundary for both backbones would miss these backbone-specific performance regions.

These sweeps show that \textsc{Strap} does not rely on a manually tuned downstream split.
Instead, the sensitivity-derived boundary provides a strong training-free default that lies close to the empirical performance peak across adapters, backbones, and datasets.

\begin{table}[ht!]
\centering
\caption{
Ablation of spatial--temporal adapter allocation across layers for ST-Adapter.
We vary the number of spatial~(S) and temporal~(T) adapters under varying depth (parameter count varies accordingly).
Bold indicates the best result per column.
}
\label{tab:placement_ablation_st}
\small
\renewcommand{\arraystretch}{1.2}
\setlength{\tabcolsep}{5pt}
\begin{tabular}{clcccc}
\toprule
\textbf{Backbone} & \textbf{Strategy} & \textbf{Params (M)}
  & \textbf{Drive\&Act}
  & \textbf{IKEA-ASM} & \textbf{HRI-30} \\
\midrule
\multirow{7}{*}{\shortstack[l]{DINOv2\\(ST-Adapter)}}
  & \textsc{All Spatial}                            & 1.1 & 73.7 & 70.2 & 68.5 \\
  & 10S/2T                                          & 2.2 & 80.6 & 75.9 & 80.6 \\
  & 8S/4T                        & 3.1 & 81.1 & 77.1 & 87.9 \\
  & \cellcolor{gray!10}\textbf{6S/6T}$^\dagger$     & \cellcolor{gray!10}4.2 & \cellcolor{gray!10}81.9 & \cellcolor{gray!10}\textbf{77.6} & \cellcolor{gray!10}\textbf{89.2} \\
  & 4S/8T                        & 5.2 & 80.9 & 76.5 & 88.0 \\
  & 2S/10T                                          & 6.1 & \textbf{82.6} & 75.3 & 91.1 \\
  & \textsc{All Temporal}                           & 7.3 & 75.2 & 70.5 & 85.5 \\
\midrule
\multirow{7}{*}{\shortstack[l]{CLIP\\(ST-Adapter)}}
  & \textsc{All Spatial}                            & 1.1 & 73.4 & 59.3 & 66.9 \\
  & 10S/2T                                          & 2.2 & 79.9 & 68.8 & 67.5 \\
  & 8S/4T                        & 3.1 & 80.3 & 71.0 & 74.8 \\
  & 6S/6T                        & 4.2 & 80.1 & 69.2 & 79.5 \\
  & \cellcolor{gray!10}\textbf{4S/8T}$^\dagger$     & \cellcolor{gray!10}5.2 & \cellcolor{gray!10}\textbf{80.8} & \cellcolor{gray!10}\textbf{73.3} & \cellcolor{gray!10}\textbf{86.2} \\
  & 2S/10T                                          & 6.1 & 80.5 & 66.8 & 82.3 \\
  & \textsc{All Temporal}                           & 7.3 & 72.6 & 70.2 & 81.1 \\
\bottomrule
\multicolumn{6}{l}{\scriptsize $^\dagger$ STRAP-predicted boundary.}
\end{tabular}
\end{table}

\begin{table}[ht!]
\centering
\caption{
Ablation of spatial--temporal adapter allocation across layers for Frame2Freq.
We vary the number of spatial~(S) and temporal~(T) adapters under varying depth (parameter count varies accordingly).
Bold indicates the best result per column.
}
\label{tab:placement_ablation_f2f}
\small
\renewcommand{\arraystretch}{1.2}
\setlength{\tabcolsep}{5pt}
\begin{tabular}{clcccc}
\toprule
\textbf{Backbone} & \textbf{Strategy} & \textbf{Params (M)}
  & \textbf{Drive\&Act}
  & \textbf{IKEA-ASM} & \textbf{HRI-30} \\
\midrule
\multirow{7}{*}{\shortstack[l]{DINOv2\\(Frame2Freq)}}
  & \textsc{All Spatial}                            & 1.1 & 73.7 & 70.2 & 68.5 \\
  & 10S/2T                                          & 2.2 & 82.4 & 77.7 & 90.1 \\
  & 8S/4T                        & 3.1 & 83.1 & 75.1 & 90.8 \\
  & \cellcolor{gray!10}\textbf{6S/6T}$^\dagger$     & \cellcolor{gray!10}4.2 & \cellcolor{gray!10}\textbf{83.5} & \cellcolor{gray!10}\textbf{79.8} & \cellcolor{gray!10}\textbf{91.2} \\
  & 4S/8T                        & 5.2 & 82.3 & 78.6 & 89.5 \\
  & 2S/10T                                          & 6.1 & 82.4 & 78.6 & 84.6 \\
  & \textsc{All Temporal}                           & 7.3 & 81.5 & 76.4 & 89.8 \\
\midrule
\multirow{7}{*}{\shortstack[l]{CLIP\\(Frame2Freq)}}
  & \textsc{All Spatial}                            & 1.1 & 73.4 & 59.3 & 66.9 \\
  & 10S/2T                                          & 2.2 & 80.7 & 68.4 & 71.4 \\
  & 8S/4T                        & 3.1 & 80.8 & 72.8 & 76.4 \\
  & 6S/6T                        & 4.2 & 81.0 & 74.2 & 83.4 \\
  & \cellcolor{gray!10}\textbf{4S/8T}$^\dagger$     & \cellcolor{gray!10}5.2 & \cellcolor{gray!10}\textbf{81.5} & \cellcolor{gray!10}76.5 & \cellcolor{gray!10}\textbf{86.3} \\
  & 2S/10T                                          & 6.1 & 81.4 & 72.1 & 82.1 \\
  & \textsc{All Temporal}                           & 7.3 & 77.4 & \textbf{78.0} & 85.7 \\
\bottomrule
\multicolumn{6}{l}{\scriptsize $^\dagger$ STRAP-predicted boundary.}
\end{tabular}
\end{table}

\begin{table}[ht!]
\centering
\caption{
Ablation of spatial--temporal adapter allocation across layers for M2-CLIP.
We vary the number of spatial~(S) and temporal~(T) adapters under varying depth (parameter count varies accordingly).
Bold indicates the best result per column.
}
\label{tab:placement_ablation_m2clip}
\small
\renewcommand{\arraystretch}{1.2}
\setlength{\tabcolsep}{5pt}
\begin{tabular}{clcccc}
\toprule
\textbf{Backbone} & \textbf{Strategy} & \textbf{Params (M)}
  & \textbf{Drive\&Act}
  & \textbf{IKEA-ASM} & \textbf{HRI-30} \\
\midrule
\multirow{7}{*}{\shortstack[l]{DINOv2\\(M2-CLIP)}}
  & \textsc{All Spatial}                            & 1.1  & 73.7 & 70.2 & 68.5 \\
  & 10S/2T                                          & 3.4  & 81.8 & 76.9 & 88.2 \\
  & 8S/4T                        & 5.6  & 82.5 & 78.7 & 91.0 \\
  & \cellcolor{gray!10}\textbf{6S/6T}$^\dagger$     & \cellcolor{gray!10}7.7  & \cellcolor{gray!10}\textbf{82.9} & \cellcolor{gray!10}\textbf{79.6} & \cellcolor{gray!10}\textbf{92.3} \\
  & 4S/8T                        & 9.9  & 82.3 & 79.1 & 90.2 \\
  & 2S/10T                                          & 12.1 & 82.5 & 77.4 & 91.6 \\
  & \textsc{All Temporal}                           & 14.8 & 77.7 & 74.9 & 80.5 \\
\midrule
\multirow{7}{*}{\shortstack[l]{CLIP\\(M2-CLIP)}}
  & \textsc{All Spatial}                            & 1.1  & 73.4 & 59.3 & 66.9 \\
  & 10S/2T                                          & 3.4  & 81.4 & 70.6 & 84.4 \\
  & 8S/4T                        & 5.6  & 81.9 & 74.5 & 91.5 \\
  & 6S/6T                        & 7.7  & 82.1 & 75.6 & 92.2 \\
  & \cellcolor{gray!10}\textbf{4S/8T}$^\dagger$     & \cellcolor{gray!10}9.9  & \cellcolor{gray!10}\textbf{82.9} & \cellcolor{gray!10}\textbf{76.2} & \cellcolor{gray!10}\textbf{93.1} \\
  & 2S/10T                                          & 12.1 & 82.1 & 74.6 & 92.8 \\
  & \textsc{All Temporal}                           & 14.8 & 77.2 & 71.9 & 85.9 \\
\bottomrule
\multicolumn{6}{l}{\scriptsize $^\dagger$ STRAP-predicted boundary.}
\end{tabular}
\end{table}

\subsection{Few-Shot Action Recognition}
\label{app:fewshot}

\noindent\textbf{Datasets.}
We evaluate few-shot generalization on SSv2-Small~\cite{ssv2}, HMDB-51~\cite{hmdb51}, and UCF-101~\cite{ucf101} under 1-shot and 5-shot protocols, following standard few-shot video benchmarks~\cite{D2_ST,trokens}. 
SSv2-Small emphasizes motion-dependent temporal reasoning, while HMDB-51 and UCF-101 cover broader human actions with varying reliance on temporal structure. 
%Together, these datasets test whether the sensitivity-derived adapter placement remains useful when only very limited labeled data are available for downstream adaptation.

\noindent\textbf{Results.}
Table~\ref{tab:fewshot} reports Top-1 accuracy for ST-Adapter, Frame2Freq, and M2-CLIP under uniform and \textsc{Strap} placement using a CLIP backbone. 
\textsc{Strap} improves over uniform placement in all 18 few-shot settings, with gains ranging from $+0.2\%$  for Frame2Freq on SSv2-Small and HMDB-51 in the 5-shot setting to $+2.0\%$  for Frame2Freq on HMDB-51 in the 1-shot setting. 
%At the same time, \textsc{Strap} reduces trainable parameters by 29--33\%. 
These consistent gains across adapter designs, datasets, and shot levels suggest that the sensitivity-derived boundary is not a dataset-specific artifact. 
Instead, it provides a useful placement prior even in low-data regimes, where inefficient or poorly placed adaptation can more easily overfit.

\begin{table*}[ht!]
\centering
\scriptsize
\setlength{\tabcolsep}{3pt}
\renewcommand{\arraystretch}{1.2}
\caption{\textbf{Few-shot action recognition.} 
\textsc{Strap} improves over uniform placement in all 18 settings with no degradation, while reducing parameters by 29-33\%. State-of-the-art methods are shown for reference; \textsc{Strap} achieves competitive performance with substantially fewer parameters.}
\label{tab:fewshot}
\begin{tabular}{lc ccc cc cc}
\toprule
& & &
\multicolumn{2}{c}{\textbf{SSv2-Small}} &
\multicolumn{2}{c}{\textbf{HMDB-51}} &
\multicolumn{2}{c}{\textbf{UCF-101}} \\
\cmidrule(lr){4-5} \cmidrule(lr){6-7} \cmidrule(lr){8-9}
\textbf{Method} & \textbf{Strategy} &
  \textbf{Params(M)} &
\textbf{1-shot} & \textbf{5-shot} &
\textbf{1-shot} & \textbf{5-shot} &
\textbf{1-shot} & \textbf{5-shot} \\
\midrule

% SOTA methods - grayed
\multicolumn{8}{l}{\textit{\textcolor{gray}{\textbf{State-of-the-Art Methods}}}} \\
\midrule
\textcolor{gray}{TATs~\cite{tats}} 
& \textcolor{gray} {-}
&\textcolor{gray}{18.1}
& \textcolor{gray}{47.9} & \textcolor{gray}{64.4} 
& \textcolor{gray}{60.0} & \textcolor{gray}{77.0} 
& \textcolor{gray}{92.0} & \textcolor{gray}{95.5} \\
\textcolor{gray}{Trokens~\cite{trokens}} 
& \textcolor{gray}{-} 
& \textcolor{gray}{17.4}
& \textcolor{gray}{53.4} & \textcolor{gray}{68.9} 
& \textcolor{gray}{69.8} & \textcolor{gray}{82.3} 
& \textcolor{gray}{94.0} & \textcolor{gray}{97.9} \\
\textcolor{gray}{D2-ST Adapter~\cite{D2_ST}} 
& \textcolor{gray}{-} 
& \textcolor{gray}{7.3}
& \textcolor{gray}{55.0} & \textcolor{gray}{69.3} 
& \textcolor{gray}{77.1} & \textcolor{gray}{88.2} 
& \textcolor{gray}{96.4} & \textcolor{gray}{99.1} \\

\midrule
% Dashed separator
\multicolumn{8}{l}{\textit{\textbf{Parameter Efficient Fine-tuning Frameworks}}} \\
\midrule

\rowcolor{gray!10}
 & Uniform
 & 7.3
 & 53.1 & 68.0
 & 74.1 & 87.3
 & 95.7 & 98.7 \\
\rowcolor{gray!10}\multirow{-2}{*}{ST-Adapter~\cite{St_adaptor}}
 & \textsc{Strap}
 & \textbf{5.2}\dpospar{29\%}
 & \textbf{53.6}\dpos{0.5} & \textbf{69.1}\dpos{1.1}
 & \textbf{74.9}\dpos{0.8} & \textbf{87.8}\dpos{0.5}
 & \textbf{96.0}\dpos{0.3} & \textbf{99.2}\dpos{0.5} \\

\multirow{2}{*}{Frame2Freq~\cite{frame2freq}}
 & Uniform
 & 7.3
 & 55.7 & 71.3
 & 71.6 & 85.5
 & 93.6 & 98.6 \\
 & \textsc{Strap}
 & \textbf{5.2}\dpospar{29\%}
 & \textbf{56.0}\dpos{0.3} & \textbf{71.5}\dpos{0.2}
 & \textbf{73.6}\dpos{2.0} & \textbf{85.7}\dpos{0.2}
 & \textbf{95.3}\dpos{1.7} & \textbf{99.2}\dpos{0.6} \\

\rowcolor{gray!10}
 & Uniform
 & 14.8
 & 54.1 & 70.7
 & 73.6 & 87.4
 & 94.9 & 99.0 \\
\rowcolor{gray!10}\multirow{-2}{*}{M2-CLIP~\cite{m2_clip}}
 & \textsc{Strap}
 & \textbf{9.9}\dpospar{33\%}
 & \textbf{54.8}\dpos{0.7} & \textbf{71.1}\dpos{0.4}
 & \textbf{74.4}\dpos{0.8} & \textbf{87.7}\dpos{0.3}
 & \textbf{96.1}\dpos{1.2} & \textbf{99.4}\dpos{0.4} \\

\bottomrule
\end{tabular}
\end{table*}

\begin{table}[ht!]
\centering
\caption{
    \textbf{Fine-grained action recognition with symmetric and non-symmetric breakdown.}
    Sym and N-Sym denote accuracy on symmetric and non-symmetric action subsets, respectively; Top-1 is overall accuracy. STRAP's largest gains concentrate on symmetric actions, which require distinguishing temporal order and direction -precisely the structure
    targeted by late-layer temporal adapters.
}
\label{tab:sym_nonsym}
\small
\setlength{\tabcolsep}{4pt}
\renewcommand{\arraystretch}{1}
\begin{tabular}{lc ccc | ccc |ccc}
\toprule
& & \multicolumn{3}{c}{\textbf{Drive\&Act}} & \multicolumn{3}{c}{\textbf{IKEA-ASM}} & \multicolumn{3}{c}{\textbf{HRI-30}} \\
\cmidrule(lr){3-5} \cmidrule(lr){6-8} \cmidrule(lr){9-11}
\textbf{Method} & \textbf{Strategy} & Sym & N-Sym & Top-1 & Sym & N-Sym & Top-1 & Sym & N-Sym & Top-1 \\
\midrule
\multicolumn{11}{l}{\textit{\textbf{DINOv2-B/14 backbone (6S/6T, derived from VideoMAE})}} \\
\midrule
\multirow{2}{*}{ST-Adapter \cite{St_adaptor}}  & Uniform        & 66.4          & 83.4          & 75.2          & 68.5          & 73.9          & 70.5          & 81.5          & 95.3          & 85.5          \\
                              & \textsc{Strap} & \textbf{78.0} & \textbf{85.5} & \textbf{81.9} & \textbf{80.3} & \textbf{76.2} & \textbf{77.6} & \textbf{85.4} & \textbf{96.8} & \textbf{89.2} \\
\addlinespace
\multirow{2}{*}{Frame2Freq \cite{frame2freq}}  & Uniform        & 77.1          & 85.5          & 81.5          & 81.0          & 74.2          & 76.4          & 85.6          & \textbf{98.9} & 89.8          \\
                              & \textsc{Strap} & \textbf{79.6} & \textbf{87.1} & \textbf{83.5} & \textbf{86.8} & \textbf{76.4} & \textbf{79.8} & \textbf{88.2} & 97.2          & \textbf{91.2} \\
\addlinespace
\multirow{2}{*}{M2-CLIP \cite{m2_clip}}     & Uniform        & 68.4          & 85.0          & 77.7          & 75.3          & 74.6          & 74.9          & 78.4          & 84.6          & 80.5          \\
                              & \textsc{Strap} & \textbf{78.8} & \textbf{86.6} & \textbf{82.9} & \textbf{83.7} & \textbf{77.6} & \textbf{79.6} & \textbf{89.3} & \textbf{98.2} & \textbf{92.3} \\
\midrule
\multicolumn{11}{l}{\textit{\textbf{CLIP-B/16 backbone (4S/8T, derived from ViCLIP)}}} \\
\midrule
\multirow{2}{*}{ST-Adapter \cite{St_adaptor}}  & Uniform        & 55.5          & \textbf{88.4} & 72.6          & 70.3          & 70.7 & 70.4          & 73.8          & 95.7          & 81.1          \\
                              & \textsc{Strap} & \textbf{73.5} & 87.3          & \textbf{80.8} & \textbf{80.0} & \textbf{70.8 }         & \textbf{73.3} & \textbf{80.0} & \textbf{98.5} & \textbf{86.2} \\
\addlinespace
\multirow{2}{*}{Frame2Freq \cite{frame2freq}}  & Uniform        & 64.4          & \textbf{88.6} & 77.4          & \textbf{83.9} & \textbf{75.1} & \textbf{78.0} & 79.1          & \textbf{98.9} & 85.7          \\
                              & \textsc{Strap} & \textbf{75.1} & 87.4          & \textbf{81.5} & 83.3          & 73.8          & 76.5          & \textbf{80.4} & 98.2          & \textbf{86.3} \\
\addlinespace
\multirow{2}{*}{M2-CLIP \cite{m2_clip}}     & Uniform        & 67.4          & 85.7          & 77.2          & 76.2          & 72.6          & 71.9          & 81.8          & 95.1          & 85.9          \\
                              & \textsc{Strap} & \textbf{76.9} & \textbf{88.4} & \textbf{82.9} & \textbf{82.5} & \textbf{73.2} & \textbf{76.2} & \textbf{90.2} & \textbf{98.9} & \textbf{93.1} \\
\bottomrule
\end{tabular}
\end{table}
\subsection{Symmetric Action Recognition}
\label{app:sym_nonsym}

Table~\ref{tab:sym_nonsym} extends the symmetric and non-symmetric action breakdown from the main paper by including CLIP-B/16 results in addition to DINOv2-B/14.

The trend observed with DINOv2 also holds for CLIP: \textsc{Strap} produces its largest gains on symmetric action subsets, where classes share similar appearance but differ in temporal order or direction.
For CLIP with ST-Adapter, symmetric accuracy improves by +18.0\% on Drive\&Act (55.5$\rightarrow$73.5) and +9.7\% on IKEA-ASM (70.3$\rightarrow$80.0).
For CLIP with M2-CLIP, symmetric accuracy improves by +9.5\% on Drive\&Act and +6.3\% on IKEA-ASM.
These gains are generally larger than those on non-symmetric subsets, where appearance cues are more discriminative.

There are a few exceptions. With CLIP and Frame2Freq, \textsc{Strap} improves symmetric accuracy on Drive\&Act and HRI-30, but drops on IKEA-ASM, matching the overall accuracy drop for this adapter--backbone pair in Table~\ref{tab:finegrained}.
With CLIP and ST-Adapter, non-symmetric accuracy on Drive\&Act decreases slightly, suggesting that the placement shift can redistribute capacity toward temporal disambiguation at a small cost to appearance-driven classes.

Overall, the symmetric-action results support the main claim of \textsc{Strap}: concentrating temporal adapters in later layers improves recognition most when temporal order is class-discriminative.
\subsection{Temporal Action Detection on THUMOS14}
\label{app:tad}

To test whether the STRAP boundary transfers beyond action recognition, we
apply it to temporal action detection on THUMOS14 using the OpenTAD \cite{opentad} framework with a VideoMAE \cite{tong2022videomae} backbone at the standard 768-frame input. We use the derived boundary directly, without any new sensitivity analysis, and train all three configurations under identical settings. We use a video-pretrained backbone rather than an image backbone, which would have required additional hyperparameter tuning. STRAP achieves a higher average mAP than uniform placement with fewer parameters, and is ahead at 4 of 5 tIoU thresholds. Gains are small, which is consistent with THUMOS14 not being a fine-grained benchmark (Sec. \ref{sec:generalization}).

\begin{table}[ht!]
\centering
\small
\setlength{\tabcolsep}{4pt}
\caption{\textbf{Temporal action detection on THUMOS14} (VideoMAE, OpenTAD).
Average mAP and mAP at each tIoU threshold. Best result per accuracy column
in bold.}
\label{tab:thumos}
\begin{tabular}{lcccccc c}
\toprule
 & & \multicolumn{5}{c}{mAP @ tIoU} & \\
\cmidrule(lr){3-7}
Config & Avg & 0.3 & 0.4 & 0.5 & 0.6 & 0.7 & Params \\
\midrule
Spatial-only \cite{adaptformer} & 69.03 & 84.76 & 79.93 & 72.34 & 61.53 & 46.62 & 1.2M \\
Uniform \cite{AdaTAD}      & 70.36 & 85.58 & \textbf{81.32} & 73.96 & 61.95 & 49.00 & 4.0M \\
\textsc{Strap}        & \textbf{70.75} & \textbf{86.06} & 80.88 & \textbf{74.41} & \textbf{63.05} & \textbf{49.34} & 3.1M \\
\bottomrule
\end{tabular}
\end{table}